\documentclass[journal]{IEEEtran}
\usepackage{xcolor,soul,framed} 
\colorlet{shadecolor}{yellow}
\usepackage{graphicx}
\graphicspath{{../pdf/}{../jpeg/}}
\usepackage[cmex10]{amsmath}
\usepackage{array}
\usepackage{mdwmath}
\usepackage{mdwtab}
\usepackage{eqparbox}
\usepackage{url}
\usepackage{bm}
\usepackage{enumerate}
\usepackage{amsfonts}
\usepackage{nomencl}
\usepackage{etoolbox}
\usepackage{amssymb}
\usepackage{mathrsfs}
\usepackage{amsmath}
\usepackage{amsfonts,amssymb}
\usepackage{float}
\graphicspath{{figures/}}
\usepackage{calc}
\usepackage{tabularx}
\usepackage{fancybox}
\usepackage{subfig}
\usepackage{xcolor}
\usepackage[ruled,linesnumbered,vlined]{algorithm2e}
\usepackage{bm}
\usepackage{bbm}

\usepackage{amsmath} % 在导言区
\allowdisplaybreaks[3] %

\usepackage{pdfpages}
\usepackage{svg}
\usepackage{booktabs}
\usepackage{multirow}
\usepackage{makecell}
\usepackage{array}
\usepackage{hyperref}
\usepackage{graphicx}
\usepackage{subcaption}
\usepackage[font=footnotesize]{caption}
\begin{document}

\title{Decision-Focused Scenario Generation and Selection for Efficient and Robust Grid Dispatch}

%Title 2
%Decision-Oriented Scenario Generation and Selection for Efficient and Robust Grid Dispatch

\author{Yangze Zhou,  Yihong Zhou, Thomas Morstyn, and Yi Wang
% \thanks{The work was supported in part by the Research Grants Council of the Hong Kong SAR  under Grant HKU 27203723 and in part by in part by the National Natural Science Foundation of China under Grant 72131001. (\textit{Corresponding authors: Yi Wang}) }
% \thanks{Yangze Zhou, and Yi Wang are with the Department of Electrical and Computer Engineering, The University of Hong Kong, Hong Kong SAR, China (e-mail: yzzhou@connect.hku.hk, yiwang@eee.hku.hk).}
% \thanks{Daniel Kirschen, is with Department of Electrical and Computer Engineering, University of Washington, Seattle, 98105, USA.(e-mail: kirschen@uw.edu, thomas.morstyn@eng.ox.ac.uk).}
% \thanks{Yihong Zhou, and Thomas Morstyn are with Department of Engineering Science, University of Oxford, OX1 3PJ Oxford, U.K.(e-mail: yihong.zhou@eng.ox.ac.uk, thomas.morstyn@eng.ox.ac.uk).}
}

% The paper headers
\markboth{Submitted to IEEE Trans. on Power Systems}%
{Shell \MakeLowercase{\textit{et al.}}: Bare Demo of IEEEtran.cls for IEEE Journals}
\maketitle

\begin{abstract}
% The growing variability caused by flexible demand and renewable generation has made uncertainty modeling increasingly important for power system dispatch, driving decision-making from deterministic paradigms toward uncertainty-aware frameworks such as distributionally robust optimization (DRO). 
The increasing uncertainty from flexible demand and renewable generation has made distributionally robust optimization (DRO) an important tool for robust power system dispatch. DRO relies on forecast scenarios to construct ambiguity sets, but conventional scenario generation pipelines are often trained in an accuracy-oriented manner and may neglect spatial correlations among uncertainties. This mismatch can produce ambiguity sets that are statistically plausible but suboptimal for downstream operation. This work proposes a decision-focused generative framework for correlated scenario generation in DRO-based dispatch. Instead of training generative models solely to fit the historical uncertainty distribution, the proposed framework optimizes generated scenarios according to their induced downstream operational cost. The proposed framework is tailored to mainstream generative models, including variational autoencoders, generative adversarial networks, and diffusion models, while capturing the joint distribution of uncertainties across buses. To improve computational tractability, we further develop a differentiable scenario selector that selects decision-relevant scenarios from a generated pool and can be trained within the same decision-focused pipeline. Case studies demonstrate that the proposed framework effectively reduces 0.80\%--2.02\% operational cost across different generative models compared to accuracy-oriented methods.

%Variational Auto-Encoder (VAE), Generative Adversarial Network (GAN), and diffusion models. 
\end{abstract}

\IEEEpeerreviewmaketitle

% 新能源和负荷都有相关性
% correlated
% 相依性
% dependent 和 correlated

\section{Introduction}
\IEEEPARstart{W}{ith} the increasing penetration of flexible loads and renewable energy sources, energy generation and demand exhibit stronger variability and randomness \cite{li2021optimal}. %perera2020quantifying
Consequently, the uncertainties associated with energy generation and consumption have become a critical factor affecting the security and economic efficiency of power system operations.

Traditional deterministic decision-making approaches are often inadequate when system conditions are subject to substantial uncertainty. To explicitly account for uncertain variations, decision-making under uncertainty (DUU) methods have been developed, among which stochastic optimization (SO) \cite{zheng2014stochastic} and robust optimization (RO) \cite{chen2022robust} are two representative approaches. SO optimizes operational decisions under an assumed probability distribution, but its out-of-sample performance may deteriorate when the adopted distribution deviates from the true one. In contrast, RO does not require an exact probability model and guarantees performance against all realizations within a prescribed uncertainty set, but this protection is often achieved at the expense of excessive conservativeness \cite{zhao2019two}. To better balance economic efficiency and robustness, distributionally robust optimization (DRO) has attracted increasing attention in recent years \cite{li2023data}. Rather than relying on a single estimated distribution or guarding against the worst-case realization, DRO optimizes against the worst-case distribution within a pre-specified ambiguity set.

DUU-based dispatch frameworks typically require various scenarios as inputs, which mainly come from the probabilistic forecasts (directly using historical scenarios can also be viewed as naïve forecasting).
A substantial body of research has been devoted to developing probabilistic energy forecasting methods \cite{hong2016probabilistic}.
Traditionally, probabilistic energy forecasting models are trained in an accuracy-oriented manner, meaning that the learning objective is to minimize forecasting errors, typically measured by metrics such as the pinball loss.
However, probabilistic energy forecasting ultimately serves downstream DUU-based system operation, and minimizing forecast errors does not necessarily translate into lower operational costs. 
This is because forecasting errors enter system operation through dispatch decisions, and under-forecasting and over-forecasting typically incur different corrective actions and penalties. %\cite{zhou2024load}.
% Specifically, under-forecasting and over-forecasting can affect operation asymmetrically: the former may trigger emergency unit commitment or expensive external power purchases, whereas the latter can lead to inefficient scheduling and wasted resources \cite{zhou2024load}.
This mismatch motivates decision-focused forecasting approaches that explicitly account for the downstream cost in model training \cite{zhang2025decision}.
For instance, \cite{han2021task} proposed decision-focused day-ahead (DA) load forecasting methods for SO-based system dispatch.
\cite{zhao2021cost} achieved decision-focused interval predictions, where the upper and lower quantile bounds are determined by minimizing the costs of an SO problem.
\cite{zhang2024toward} developed a decision-focused scenario generation method for sequential energy dispatch iteratively.
\cite{carriere2019integrated} coupled quantile-based forecasting of solar and wind generation with day-ahead market bidding
%, formulating the bidding task as an SO problem 
for revenue maximization. \cite{stratigakos2025decision} combined multiple probabilistic energy forecasting models using an ensemble scheme trained to minimize the expected decision cost associated with an SO problem. Similar works can also be found in \cite{huang2025advancing,zhuang2025weighted}.
However, these studies share a common limitation: \textit{they focus on modeling the marginal distribution of time series independently, thereby overlooking the multivariate correlation between them.}

% This limitation is largely rooted in the difficulty of characterizing the high-dimensional joint distribution of uncertainty using traditional probabilistic energy forecasting approaches. 
This limitation reflects a broader challenge: accurately characterizing the high-dimensional joint distribution of uncertainty remains difficult for traditional probabilistic energy forecasting approaches, particularly when the generated forecasts are used for downstream decision-making.
In parametric methods, probabilistic forecasts are typically represented through a predefined distribution family with a limited number of parameters, such as the mean and variance \cite{hong2016probabilistic}. Extending such formulations to jointly capture spatial and temporal correlations often requires specifying a high-dimensional multivariate distribution and estimating a large covariance structure, which can be computationally demanding and statistically unreliable \cite{salinas2019high}. Non-parametric methods, on the other hand, usually generate quantile estimation or prediction intervals for individual variables. 
To introduce correlations, some studies combine these marginal forecasts with copula models to generate correlated scenarios \cite{panamtash2020copula}. However, their effectiveness is often limited in high-dimensional settings, where the specification of copula models and estimation of associated parameters become increasingly difficult \cite{salinas2019high}. 
As a result, although correlated probabilistic forecasting has been considered, traditional probabilistic approaches still face challenges in accurately representing the full joint spatial uncertainty, especially in high-dimensional settings.

These challenges motivate the use of generative models, which have attracted growing interest in probabilistic energy forecasting because of their ability to learn complex high-dimensional distributions directly from data without imposing strong parametric assumptions.
% In this context, generative models have attracted growing interest in probabilistic energy forecasting because of their ability to learn complex high-dimensional distributions directly from data without imposing strong parametric assumptions. 
From a modeling perspective, existing generative methods can generally be categorized into three paradigms: explicit density models (e.g., VAEs \cite{he2024distributional}), implicit density models (e.g., GANs \cite{wang2023operational}), and continuous-time generative models (e.g., diffusion models \cite{li2024transformer}) \cite{mehmood2023deep}. 
% For instance, \cite{wang2023operational} employed a Wasserstein generative adversarial network (WGAN) to model multivariate load uncertainty without relying on an explicit distributional form. \cite{he2024distributional} proposed a temporal conditional variational autoencoder (TCVAE) to capture the time-varying joint probability distribution of multivariate time series under distribution drift. \cite{li2024transformer} introduced a Transformer-Modulated Diffusion Model (TMDM), where transformers provide historical dependency information to guide a conditional diffusion process for uncertainty-aware multivariate time series forecasting.
%For example, \cite{wang2023operational} adopted a Wasserstein GAN for multivariate load uncertainty modeling, \cite{he2024distributional} developed a temporal conditional VAE for distributional forecasting, and \cite{li2024transformer} proposed a Transformer-Modulated Diffusion Model for uncertainty-aware multivariate time series forecasting.
%Despite these advances, existing generative forecasting models are still predominantly trained in an accuracy-oriented manner. 
Despite these advances in correlated probabilistic forecasting, existing generative probabilistic forecasting models are still predominantly trained in an accuracy-oriented manner, rather than being directly optimized for downstream operational objectives. One major obstacle to decision-focused training lies precisely in the heterogeneity of these generative paradigms. Since VAEs, GANs, and diffusion models differ fundamentally in their data-generation mechanisms and optimization objectives, training strategies tailored to one class are generally difficult to extend to others. As a result, \textit{a unified decision-focused framework for generative probabilistic forecasting in downstream system operation has yet to be well established.}
%As a result, \textit{a unified decision-focused framework for generative models in downstream system operation has yet to be well established.} 

Training a decision-focused forecasting model is often computationally intensive, as it typically involves repeatedly solving downstream optimization problems and computing the feedback (e.g., gradient of dispatch cost over scenarios) during the learning process.
This challenge is particularly pronounced for DRO-based operation, where even a single optimization run can be time-consuming if the scenario set is huge \cite{zhou2025fica}. As a result, scenario selection becomes a natural strategy for improving computational tractability. For SO problems, scenario selection has been studied, and existing methods are mainly based on statistical similarity metrics \cite{liu2017hierarchical,lin2021stochastic}. However, such methods are often non-differentiable and therefore difficult to integrate into gradient-based machine learning frameworks. As for DRO, corresponding research remains limited. 
More importantly, existing scenario selection methods are typically designed to preserve statistical characteristics of the original scenario set, rather than the downstream operational objective. As a result, a reduced scenario set with high statistical similarity may still lead to suboptimal operational decisions and additional cost when used in system operation. Although \cite{wan2026cost} proposed a cost-oriented scenario selection method, their study is limited to SO and overlooks the correlations among different sources of uncertainty. This motivates an important yet underexplored question: \textit{how can scenarios be selected according to their impact on downstream DRO decisions to enable efficient and robust dispatch?}

To this end, this paper makes the following contributions:
\begin{enumerate}
    \item We propose a unified decision-focused scenario generation framework, which integrates the generative model and systems dispatch based on DRO. The proposed framework is model-agnostic and can be instantiated with a wide range of generative paradigms, including explicit density models (VAE), implicit density models (GAN), and continuous-time generative models (diffusion), rather than being tailored to a specific architecture. 
     % \item We developed a correlated scenario generation method that learns the joint distribution of bus-level uncertainty using generative models, instead of constructing scenarios independently from marginal distributions. By preserving inter-bus correlation structures, the proposed method yields more realistic ambiguity sets for DRO and improves downstream dispatch decisions and operational cost performance.
    \item We develop a correlation-aware scenario generation strategy that constructs bus-level uncertainty scenarios from a learned joint distribution, rather than independently combining marginal forecasts. Unlike conventional scenario generation methods that mainly preserve statistical forecasting accuracy, the generated scenarios are evaluated through their impact on downstream DRO dispatch, yielding more decision-relevant ambiguity sets and improved operational cost performance.
     % \item We design a data-driven scenario selector to reduce the computational burden associated with repeated downstream optimization and gradient calculation. Compared with statistical similarity metric-based scenario selection methods, the proposed selector is differentiable and can be trained in a decision-focused approach by integrating it into gradient-based learning pipelines. The proposed scenario selection procedure is model-agnostic as well, which means it can be combined with mainstream probabilistic energy forecasting methods.
    \item We design a data-driven scenario selector to reduce the computational burden of repeated downstream optimization and gradient calculation. Unlike statistical similarity-based selection methods, the proposed selector is differentiable, decision-focused, and model-agnostic, enabling integration with mainstream probabilistic energy forecasting methods. Case studies show that the resulting end-to-end framework reduces operational cost by 0.80\%--2.02\% across different generative models compared with accuracy-oriented methods.

     % \item We evaluate our proposed framework with a real-world dataset. Our case study demonstrates that accounting for inter-nodal load correlations is crucial for reducing decision costs. This work also show that the proposed decision-focused scenario generation methods effectively reduce out-of-sample operating costs across three mainstream generative models.
\end{enumerate}

The remainder of this paper is organized as follows. Section \ref{PS} establishes the problem formulation, and Section \ref{Preliminaries} provides background on the key components of the proposed framework. Section \ref{methodology} presents the proposed decision-focused scenario generation and selection framework. Section \ref{case} describes the experimental setup and reports the results. Finally, Section \ref{conclusion} concludes the paper.

\section{Problem Statement}
\label{PS}
% Consider a network with bus set $\mathcal{B}$ and generator set $\mathcal{G}$. 
% %Let $\mathcal{B}^{\mathrm{L}} \subseteq \mathcal{B}$ and $\mathcal{B}^{\mathrm{G}} \subseteq \mathcal{B}$ denote the sets of load buses and generation buses, respectively, 
% $\mathcal{T}=\{1, ..., T\}$ collects the indices of T time steps in the scheduling horizon.
%and let $\mathcal{T}$ denote the scheduling horizon.
Consider a transmission network and let set $\mathcal{B}$, $\mathcal{G}$, and $\mathcal{T}$ collect the indices for all buses, generators, and time steps in the dispatch horizon. 
Denote $\mathcal{B}^{\mathrm{U}}\subseteq \mathcal{B} $ by the buses associated with uncertain quantities of interest, such as load, renewable energy, and net load.
For each $b \in \mathcal{B}^{\mathrm{U}}$, let $\boldsymbol{X}_b$ denote the input features associated with bus $b$, and let $\boldsymbol{y}_b=\{y_{b,t}\}_{t\in\mathcal{T}}$ denote the corresponding uncertain profile to be forecast.

DUU methods require a set of plausible future realizations of these quantities as input, referred to as scenarios. 
Existing probabilistic forecasting approaches often train a separate model for each uncertain quantity at each bus. Specifically, for each $b \in \mathcal{B}^{\mathrm{U}}$, an individual predictor $f_b$ parameterized by $\boldsymbol{w}_b$ is trained to characterize the marginal distribution of $\boldsymbol{y}_b$ (based on $\boldsymbol{X}_b$), denoted by $\mathbb{P}(\boldsymbol{y}_b)$. Collect all the parameters as $\boldsymbol{w}_{\mathrm{sep}}=\{\boldsymbol{w}_b\}_{b\in\mathcal{B}^{\mathrm{U}}}$.
Given the marginal distributions characterized by these separately trained predictors,
%Given these marginal distributions, 
scenarios are typically constructed by independently sampling $\hat{\boldsymbol{y}}_b^{(s)}\sim\mathbb{P}(\boldsymbol{y}_b)$ for all $b \in \mathcal{B}^{\mathrm{U}}$ and $s\in\{1,\dots,S\}$, and then concatenating them into a combined scenario $\hat{\boldsymbol{y}}_{\mathrm{sep}}^{(s)}=\{\hat{\boldsymbol{y}}_b^{(s)}\}_{b\in\mathcal{B}^{\mathrm{U}}}$. The resulting scenario set is denoted by $\mathcal{S}_{\mathrm{sep}}(\boldsymbol{w}_{\mathrm{sep}})=\{\hat{\boldsymbol{y}}_{\mathrm{sep}}^{(s)}\}_{s=1}^{S}$. This construction ignores inter-bus correlations and may produce jointly unrealistic uncertainty patterns, which can adversely affect downstream dispatch decisions \cite{yang2019analytical}.
Specifically, nearby buses often share common exogenous drivers, such as temperature or solar irradiance, which induce correlated load or renewable generation patterns. As a result, it is unlikely for one bus to exhibit an unusually high realization while another nearby bus exhibits an unusually low one.

To preserve inter-bus correlations, this work instead employs a joint generative forecasting model $f_{\mathrm{joint}}$ parameterized by $\boldsymbol{w}_{\mathrm{joint}}$ to characterize the joint distribution of correlated uncertainty quantities over all buses, denoted by $\mathbb{P}(\boldsymbol{y}_{\mathrm{joint}})$, where $\boldsymbol{X}_{\mathrm{joint}}=\{\boldsymbol{X}_b\}_{b\in\mathcal{B}^{\mathrm{U}}}$ and $\boldsymbol{y}_{\mathrm{joint}}=\{\boldsymbol{y}_b\}_{b\in\mathcal{B}^{\mathrm{U}}}$. The generated scenario set is written as $\mathcal{S}(\boldsymbol{w}_{\mathrm{joint}})=\{\hat{\boldsymbol{y}}_{\mathrm{joint}}^{(s)}\}_{s=1}^{S}$, where each $\hat{\boldsymbol{y}}_{\mathrm{joint}}^{(s)}$ is one realization of $\boldsymbol{y}_{\mathrm{joint}}$.

Conventional forecasting models are usually trained to minimize forecasting errors (e.g., the pinball loss \cite{hong2016probabilistic}). However, the goal of DUU-based operation is to minimize the operational cost that depends on the dispatch decision and the realized uncertainty. This mismatch motivates decision-focused learning, where the forecasting model is trained using the more direct downstream decision cost induced by the forecasting scenarios. 
Mathematically, the decision-focused learning can be expressed as:
\begin{equation}
\begin{aligned}
&\min_{\boldsymbol{w},\boldsymbol{z}}\mathcal{C}(\boldsymbol{z},\boldsymbol{y}, \mathcal{S}(\boldsymbol{w}))\quad\boldsymbol{z}\in\mathcal{Z}(\mathcal{S}(\boldsymbol{w}))
 \end{aligned}
\end{equation}
% Mathematically, decision-focused learning can be expressed as:
% \begin{equation}
% \begin{aligned}
% \min_{\boldsymbol{w}} \quad 
% & \mathcal{C}\left(\boldsymbol{z}^*(\mathcal{S}(\boldsymbol{w})), \boldsymbol{y}\right) \\
% \text{s.t.} \quad 
% & \boldsymbol{z}^*(\mathcal{S}(\boldsymbol{w})) 
% \in \arg\min_{\boldsymbol{z}\in\mathcal{Z}(\mathcal{S}(\boldsymbol{w}))}
% \mathcal{C}\left(\boldsymbol{z},\mathcal{S}(\boldsymbol{w})\right).
% \end{aligned}
% \end{equation}
Here, $\mathcal{Z}(\mathcal{S}(\boldsymbol{w}))$ denotes the feasible set defined by the downstream dispatch constraints under the scenario set $\mathcal{S}(\boldsymbol{w})$, and $\mathcal{C}(\boldsymbol{z},\boldsymbol{y},\mathcal{S}(\boldsymbol{w}))$ denotes the operational cost under decision $\boldsymbol{z}$, actual realization $\boldsymbol{y}$, and the scenario set generated by a model parameterized by $\boldsymbol{w}$. The downstream dispatch yields an optimal decision $\boldsymbol{z}^{*}(\mathcal{S}(\boldsymbol{w}))$ and the corresponding minimized cost $\mathcal{C}^{*}(\mathcal{S}(\boldsymbol{w}))$.
% Here, $\mathcal{Z}(\mathcal{S}(\boldsymbol{w}))$ denotes the feasible set defined by the downstream dispatch constraints under the scenario set $\mathcal{S}(\boldsymbol{w})$. The outer objective evaluates the realized operational cost under the actual realization $\boldsymbol{y}$, while the inner problem represents the downstream dispatch problem solved using the generated scenario set.
Training a decision-focused generative model under this formulation involves two main challenges:
\begin{itemize}
\item The scenario set $\mathcal{S}(\boldsymbol{w})$ should be generated from a high-dimensional joint distribution to preserve inter-bus correlations. However, different generative models follow distinct generation mechanisms, which require a unified decision-focused training framework that is not tied to a specific architecture. Accordingly, \textit{this work develops a model-agnostic decision-focused joint scenario generation framework.}
\item The size of $\mathcal{S}(\boldsymbol{w})$ directly affects the computational tractability of downstream optimization, as large scenario sets substantially increase the cost of repeated model training and dispatch evaluation. \textit{Accordingly, this work develops a decision-focused scenario selection method that constructs a reduced set $\mathcal{S}_{\mathrm{r}}(\boldsymbol{w})$ while preserving downstream decision quality.}
\end{itemize}

\section{Preliminaries}\label{Preliminaries}

\subsection{Power System Dispatch Model}
\label{section_3_a}

% 强调一下two stage 是pratical 的
% 如果要extend to multi-stage的
% 第二阶段只是一个model的一个process
% selector 替换一下selector
% contribution把correlated单独列出来讲
% flowing matching
% quantiles copula （intro里增加一个说明）
% intro部分的SO和RO简化一些
 
% This work considers a two-stage dispatch model adapted from \cite{lu2018security}.

To instantiate the proposed framework, we consider a representative two-stage dispatch model adapted from \cite{lu2018security}. In the day-ahead (DA) stage, the system operator determines scheduled generation and reserve capacities with forecasts. In the real-time (RT) stage, once uncertainty is realized, the operator re-dispatches units using the procured reserves to maintain system balance. The proposed framework itself is not restricted to this specific dispatch setting and can be applied to other uncertainty-aware dispatch models with more general multi-stage non-anticipative recourse structures.
The DA dispatch can be modeled as:
\begin{subequations}\label{eq:DA_DC}
\begin{alignat}{3}
\min_{\substack{P^{\rm DA},\,R^\uparrow,\,R^\downarrow,\\\theta^{\rm DA},f^{\rm DA}}}\quad
& \sum_{t\in\mathcal{T}}\sum_{g\in\mathcal{G}} c_g\, P^{\rm DA}_{g,t}
 + \sum_{t\in\mathcal{T}}\sum_{g\in\mathcal{G}}
\Big(\pi_g^{\uparrow}R^{\uparrow}_{g,t}+\pi_g^{\downarrow}R^{\downarrow}_{g,t}\Big)
&&&& \label{eq:DA_obj}\\
\text{s.t.}\quad
& P_g^{\min}\le P^{\rm DA}_{g,t}\le P_g^{\max}
\label{eq:DA_genlim}\\
& 0\le R^{\uparrow}_{g,t},\, 0\le R^{\downarrow}_{g,t} \label{eq:DA_res_nonneg}\\
& P^{\rm DA}_{g,t}+R^{\uparrow}_{g,t}\le P_g^{\max} \label{eq:DA_cap_up}\\
& P^{\rm DA}_{g,t}-R^{\downarrow}_{g,t}\ge P_g^{\min}\label{eq:DA_cap_dn}\\
& -P^{\rm ramp}_g \le P^{\rm DA}_{g,t}-P^{\rm DA}_{g,t-1}\le P^{\rm ramp}_g \label{eq:DA_ramp}\\
& \sum_{g\in\mathcal{G}_b} P^{\rm DA}_{g,t}-\hat y_{b,t}
= \sum_{\ell\in\Delta(b)} f^{\rm DA}_{\ell,t}
\label{eq:DA_balance}\\
& f^{\rm DA}_{\ell,t}=B_\ell\big(\theta^{\rm DA}_{i(\ell),t}-\theta^{\rm DA}_{j(\ell),t}\big)
\label{eq:DA_dcflow}\\
& -\overline F_\ell \le f^{\rm DA}_{\ell,t}\le \overline F_\ell
 \label{eq:DA_linecap}
% & \theta^{\rm DA}_{b_0,t}=0 \label{eq:DA_ref}
\end{alignat}
\end{subequations}
where $P^{\rm DA}_{g,t}$, $R^{\uparrow}_{g,t}$, and $R^{\downarrow}_{g,t}$ denote the DA scheduled generation and the procured up-/down-reserve capacities of unit $g$ at time $t$, respectively, while $\theta^{\rm DA}_{b,t}$ and $f^{\rm DA}_{\ell,t}$ denote the bus voltage angles and line flows under the DC power-flow approximation. $\hat y_{b,t}$ denotes the forecasted realization of the uncertain quantity at bus $b$ and time $t$. $\mathcal{G}_b$ is the set of generators at bus $b$, $\Delta(b)$ is the set of lines connected to bus $b$, and $i(\ell)$ and $j(\ell)$ are the terminal buses of line $\ell$. $P_g^{\min}$, $P_g^{\max}$, and $P^{\rm ramp}_g$ denote the generation limits and ramp-rate limit, while $B_\ell$ and $\overline{F}_\ell$ denote the line susceptance and flow limit. $c_g$ is the DA energy cost coefficient, and $\pi_g^{\uparrow}$ and $\pi_g^{\downarrow}$ are the up-/down-reserve capacity prices.

The RT dispatch problem under a realized uncertainty scenario can be written as:

\begin{subequations}\label{eq:RT_DC}
\begin{alignat}{9}
\min_{\substack{d^\uparrow,\,d^\downarrow,f^{\rm RT},\\l^{\text{LS}},l^{\text{SP}},\,\theta^{\rm RT}}}
 &\sum_{t\in\mathcal{T}}\sum_{g\in\mathcal{G}}
\Big(\rho_g^{\uparrow} d^{\uparrow}_{g,t,s}+\rho_g^{\downarrow} d^{\downarrow}_{g,t,s}\Big)
&&&& \nonumber\\
&\quad + \sum_{t\in\mathcal{T}}\sum_{b\in\mathcal{B}}
\Big(\beta^{\rm shed} l^{\text{LS}}_{b,t,s}+\beta^{\rm cur} l^{\text{SP}}_{b,t,s}\Big)
&&&& \label{eq:RT_obj}\\
\textbf{s.t.}\quad
& 0\le d^{\uparrow}_{g,t,s}\le R^{\uparrow}_{g,t}\quad 0\le d^{\downarrow}_{g,t,s}\le R^{\downarrow}_{g,t}
\label{eq:RT_dn_reslink}\\
& P^{\rm RT}_{g,t,s}=P^{\rm DA}_{g,t}+d^{\uparrow}_{g,t,s}-d^{\downarrow}_{g,t,s}
 \label{eq:RT_prt_def}\\
& P_g^{\min}\le P^{\rm RT}_{g,t,s}\le P_g^{\max}
 \label{eq:RT_genlim}\\
& -P^{\rm ramp}_g \le P^{\rm RT}_{g,t,s}-P^{\rm RT}_{g,t-1,s}\le P^{\rm ramp}_g
\label{eq:RT_ramp}\\
& 0\le l^{\text{LS}}_{b,t,s}\le y_{b,t,s}\quad 0\le l^{\text{SP}}_{b,t,s}
\label{eq:RT_sp}\\
& \sum_{g\in\mathcal{G}_b} P^{\rm RT}_{g,t,s}
-\big(y_{b,t,s}-l^{\text{LS}}_{b,t,s}\big)-l^{\text{SP}}_{b,t,s}=\sum_{\ell\in\delta(b)} f^{\rm RT}_{\ell,t,s}\label{eq:RT_balance}\\
& f^{\rm RT}_{\ell,t,s}
= B_\ell\big(\theta^{\rm RT}_{i(\ell),t,s}-\theta^{\rm RT}_{j(\ell),t,s}\big)
\label{eq:RT_dcflow}\\
& -\overline F_\ell \le f^{\rm RT}_{\ell,t,s}\le \overline F_\ell
\label{eq:RT_linecap}
\end{alignat}
\end{subequations}
where $d^{\uparrow}_{g,t,s}$ and $d^{\downarrow}_{g,t,s}$ denote the deployed up-/down-reserves of unit $g$ at time $t$ in scenario $s$, and $P^{\rm RT}_{g,t,s}$ denotes the resulting RT generation. $y_{b,t,s}$ denotes the realized uncertainty at bus $b$ and time $t$ in scenario $s$, and non-dispatchable renewable generation is considered as uncertain negative demand. $l^{\rm LS}_{b,t,s}$ and $l^{\rm SP}_{b,t,s}$ denote two types of corrective actions used to restore system balance when reserves are insufficient, and $\theta^{\rm RT}_{b,t,s}$ and $f^{\rm RT}_{\ell,t,s}$ denote the RT bus voltage angles and line flows. $\rho_g^{\uparrow}$ and $\rho_g^{\downarrow}$ are the deployment costs of up-/down-reserves, while $\beta^{\rm shed}$ and $\beta^{\rm cur}$ are the penalty coefficients associated with the corresponding corrective actions.

\begin{figure*}
\centering
\includegraphics[width=0.8\linewidth]{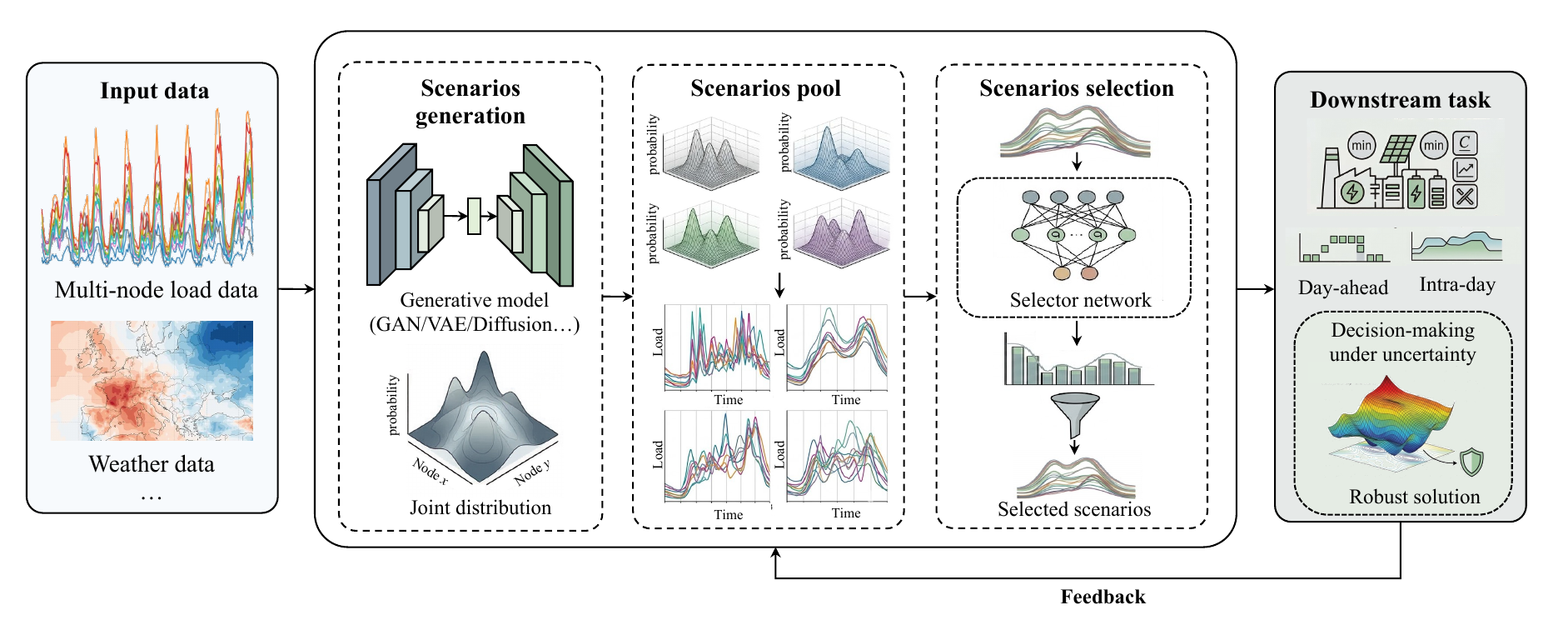}
\caption{Overview of the proposed decision-focused scenario generation framework and its interaction with downstream dispatch.}
\label{fig:framework}
\end{figure*}

\begin{figure}
\centering
\includegraphics[width=0.85\linewidth]{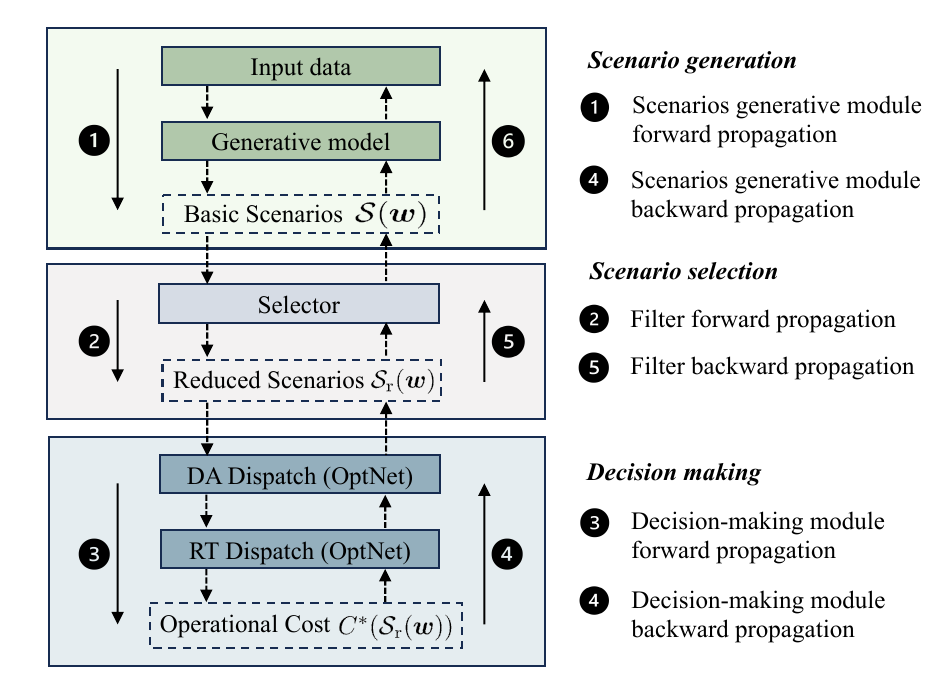}
\caption{Simplified illustration of the proposed decision-focused scenario generation framework.}
\label{fig:framework2}
\end{figure}

\subsection{DUU with DRO}

To account for uncertainty in the DA dispatch, we extend the two-stage formulation to a DRO framework. Specifically, we approximate the unknown load distribution by a distribution constructed from a finite set of data-driven scenarios. 
%Since this nominal distribution may deviate from the true out-of-sample distribution, the DRO formulation seeks decisions that are robust against such distributional ambiguity. 
Given a scenario set $\mathcal{S}(\boldsymbol{w})=\{\boldsymbol{y}^{(s)}\}_{s=1}^{S}$,
we interpret it as a data-driven proxy of the unknown load distribution and construct the nominal distribution as follows:
\begin{equation}\label{eq:Phat}
\widehat{\mathbb{P}}=\frac{1}{S}\sum_{s=1}^{S}\delta_{\boldsymbol{y}^{(s)}}
\end{equation}
where $\delta_{\boldsymbol{y}^{(s)}}$ is the Dirac measure at scenario $\boldsymbol{y}^{(s)}$. We use a unified notation
$\mathcal{S}(\boldsymbol{w})=\{\hat{\boldsymbol{y}}^{(s)}\}_{s=1}^{S} $
for the generated scenario set, where $\boldsymbol{w}$ denotes the parameters of the underlying generator. For notational simplicity, we drop the subscripts ``joint'' and ``sep'' in the remainder of this section and use $\boldsymbol{X}$ and $\boldsymbol{y}$ to denote the generator input and the actual uncertainty realization, respectively.

Since $\widehat{\mathbb{P}}$ is supported on a finite set of scenarios generated by a forecasting pipeline, and these scenarios inevitably contain forecasting errors, it may differ from the true out-of-sample distribution.
Therefore, we introduce an ambiguity set centered around $\widehat{\mathbb{P}}$ to account for such distributional uncertainty.
To remove unnecessary conservativeness, we assume uncertainty realizations lie in a known support set $\Xi$ that captures physical bounds and consistency requirements, and restrict all candidate distributions to be supported on $\Xi$. We then define the Wasserstein ambiguity set centered at $\widehat{\mathbb{P}}$ as
\begin{equation}\label{eq:Wball}
\mathbb{D}:=\Big\{\mathbb{Q}\in\mathcal{M}(\boldsymbol{\Xi}) : d_{\rm W}(\mathbb{Q},\widehat{\mathbb{P}})\le \epsilon \Big\}
\end{equation}
where $\mathcal{M}(\boldsymbol{\Xi})$ denotes the set of probability distributions supported on $\Xi$, $d_{\rm W}(\cdot,\cdot)$ is the Wasserstein distance, and $\epsilon\ge 0$ controls the robustness level.

To align with the optimization problem \eqref{eq:DA_DC} and \eqref{eq:RT_DC} in Section \ref{section_3_a}, we express the downstream dispatch under load uncertainty in an explicit two-stage Wasserstein-DRO form. 
Given the nominal distribution $\widehat{\mathbb{P}}$ and the Wasserstein ambiguity set $\mathbb{D}$, the two-stage DRO dispatch is formulated as
\begin{subequations}\label{eq:DRO_two_stage}
\begin{align}
\min_{\boldsymbol{x}} \quad &
\mathcal{C}^{\rm DA}(\boldsymbol{x})
+\sup_{\mathbb{Q}\in\mathbb{D}}
\mathbb{E}_{\boldsymbol{y}\sim\mathbb{Q}}\!\left[\Phi(\boldsymbol{x},\boldsymbol{y})\right]
\label{eq:DRO_two_stage_a}\\
\text{s.t.}\quad &
\boldsymbol{x}\ \text{satisfies}\ \eqref{eq:DA_genlim}\text{--}\eqref{eq:DA_linecap},
\label{eq:DRO_two_stage_b}\\
& \Phi(\boldsymbol{x},\boldsymbol{y})
=\min_{\boldsymbol{u}}\ \mathcal{C}^{\rm RT}(\boldsymbol{x},\boldsymbol{u},\boldsymbol{y})
\label{eq:DRO_two_stage_c}\\
& \phantom{Q(\boldsymbol{x},\boldsymbol{y})={}}\text{s.t.}\quad
(\boldsymbol{x},\boldsymbol{u},\boldsymbol{y})\ \text{satisfy}\ \eqref{eq:RT_dn_reslink}\text{--}\eqref{eq:RT_linecap}
\label{eq:DRO_two_stage_d}
\end{align}
\end{subequations}
where $\boldsymbol{x}=\{P^{\rm DA}_{g,t},R^\uparrow_{g,t},R^\downarrow_{g,t},\theta^{\rm DA}_{b,t},f^{\rm DA}_{\ell,t}\}$ is the DA decision variables, and $\boldsymbol{u}=\{d^\uparrow_{g,t},d^\downarrow_{g,t},P^{\rm RT}_{g,t},l^{\rm LS}_{b,t},l^{\rm SP}_{b,t},\theta^{\rm RT}_{b,t},f^{\rm RT}_{\ell,t}\}$ collects the RT recourse variables, for all relevant indices \(g,t,b,\ell\). Moreover, $\mathcal{C}^{\rm DA}(\boldsymbol{x})$ and $\mathcal{C}^{\rm RT}(\boldsymbol{x},\boldsymbol{u},\boldsymbol{y})$ denote the DA and RT costs in \eqref{eq:DA_obj} and \eqref{eq:RT_obj}, respectively.
% Let $
% \boldsymbol{x}=\{P^{\rm DA}_{g,t},\,R^\uparrow_{g,t},\,R^\downarrow_{g,t},\,\theta^{\rm DA}_{b,t},\,f^{\rm DA}_{\ell,t}\}_{\forall g,t,b,\ell}$
% collects the DA decisions, and let $
% \boldsymbol{u}=\{d^\uparrow_{g,t},\,d^\downarrow_{g,t},\,P^{\rm RT}_{g,t},\,l^{\rm LS}_{b,t},\,l^{\rm SP}_{b,t},\,\theta^{\rm RT}_{b,t},\,f^{\rm RT}_{\ell,t}\}_{\forall g,t,b,\ell}$
% collects the RT recourse decisions.

\section{Methodology}
\label{methodology}

The overall architecture of the proposed decision-focused scenario generation framework is illustrated in Fig.~\ref{fig:framework}. The framework consists of two main procedures: scenario generation and scenario selection.
% First, the generative model is trained using a conventional approach and then further fine-tuned in a decision-focused manner. It produces a pool of candidate scenarios, which is subsequently passed to the selecting module for scenario selection. Second, a data-driven scenario selector is trained under the same decision-focused principle to select the most informative scenarios from the candidate pool generated by the generative model. The two procedures are explained as follows.
Firstly, the generative model is trained with a decision-focused objective. It generates a pool of candidate scenarios.
Secondly, a data-driven scenario selector is trained under the same decision-focused principle to select the most informative scenarios from the candidate pool.

\subsection{Decision-focused Scenario Generation}
This section elaborates on the decision-focused training procedure for the generative model.
As shown in Fig. \ref{fig:framework2}, the proposed workflow comprises six steps, which correspond to the forward and backward propagation passes through three key components: the generative model, the scenario selection module, and the DUU module.

\textbf{Forward propagation through the generative model:} 
% To begin decision-focused training of the basic generative model, a common strategy for accelerating convergence is to first warm up the model using an accuracy-oriented objective and then fine-tune it in a decision-focused manner.
In this work, we consider three representative generative models, namely VAE, GAN, and diffusion models. 

\textit{1) VAE:}
A typical VAE consists of an encoder and a decoder, with parameters $\boldsymbol{w}=(\boldsymbol{w}_{\rm enc},\boldsymbol{w}_{\rm dec})$. Given $(\boldsymbol{X},\boldsymbol{y})$, the encoder maps $(\boldsymbol{X},\boldsymbol{y})$ to the parameters $(\boldsymbol{\mu},\boldsymbol{\sigma})$ of a Gaussian latent representation, from which the latent variable is sampled via the reparameterization trick as $\boldsymbol{r}=\boldsymbol{\mu}+\boldsymbol{\sigma}\odot\boldsymbol{\eta}$ with $\boldsymbol{\eta}\sim\mathcal{N}(\boldsymbol{0},\boldsymbol{I})$. The decoder then generates the scenario as $\boldsymbol{y}=G_{\boldsymbol{w}}(\boldsymbol{r},\boldsymbol{X})$.
The standard VAE training objective is the negative evidence lower bound (ELBO):
\begin{equation}
\begin{aligned}
\mathcal{L}_{\rm VAE}
=
&-\mathbb{E}\left[\log p_{\boldsymbol{w}_{\rm dec}}(\boldsymbol{y}\mid \boldsymbol{X},\boldsymbol{r})\right]
\\&+\mathrm{KL}\!\left(
q_{\boldsymbol{w}_{\rm enc}}(\boldsymbol{r}\mid \boldsymbol{X},\boldsymbol{y})
\,\|\, \mathcal{N}(\boldsymbol{0},\boldsymbol{I})
\right)
\end{aligned}
\end{equation}
where ${\rm KL}(\cdot\|\cdot)$ is the KL divergence.

\textit{2) GAN:}
A GAN consists of a generator $G_{\boldsymbol{w}_G}$ and a discriminator $D_{\boldsymbol{\phi}}$. Given $\boldsymbol{X}$ and a Gaussian noise vector $\boldsymbol{\eta}\sim\mathcal{N}(\boldsymbol{0},\boldsymbol{I})$, the generator produces a scenario as 
$\hat{\boldsymbol{y}}=G_{\boldsymbol{w}_G}(\boldsymbol{\eta},\boldsymbol{X})$. The discriminator $D_{\boldsymbol{\phi}}(\boldsymbol{y},\boldsymbol{X})$ aims to distinguish generated samples from real samples drawn from the training data distribution, denoted by $\mathbb{P}_{\rm data}$. The standard GAN objective is formulated as a minimax game between the generator and the discriminator:
\begin{equation}
\begin{aligned}
&\min_{\boldsymbol{w}_G}\max_{\boldsymbol{\phi}}\quad
\mathbb{E}_{\boldsymbol{y}\sim \mathbb{P}_{\rm data}}
\big[\log D_{\boldsymbol{\phi}}(\boldsymbol{y},\boldsymbol{X})\big] \\
&+\mathbb{E}_{\boldsymbol{\eta}\sim \mathcal{N}(\boldsymbol{0},\boldsymbol{I})}
\big[\log\!\big(1-D_{\boldsymbol{\phi}}(G_{\boldsymbol{w}_G}(\boldsymbol{\eta},\boldsymbol{X}),\boldsymbol{X})\big)\big]
\end{aligned}
\end{equation}

\textit{3) Diffusion:}
Among diffusion-based generative models, the denoising diffusion probabilistic model (DDPM) is one of the most widely used formulations and is adopted here as a scenario generator. Let
$\boldsymbol{h}_{0}=\boldsymbol{y}$
denote a target scenario. In the forward diffusion process, Gaussian noise is progressively added to $\boldsymbol{h}_{0}$ according to a predefined variance schedule $\{b_n\}_{n=1}^{N}$. Let $a_n=1-b_n$ and $\bar{a}_n=\prod_{i=1}^{n}a_i$. Then the perturbed state at step $n$ can be written as:
\begin{equation}
\boldsymbol{h}_{n}
=
\sqrt{\bar{a}_{n}}\ \boldsymbol{y}
+
\sqrt{1-\bar{a}_{n}}\ \boldsymbol{\eta} 
\qquad
\boldsymbol{\eta}\sim\mathcal{N}(\boldsymbol{0},\boldsymbol{I})
\end{equation}
The reverse process is parameterized by a denoising network with parameters $\boldsymbol{w}_{\rm den}$, which predicts the injected noise from $(\boldsymbol{h}_{n},n,\boldsymbol{X})$. The standard DDPM training objective is
\begin{equation}
\mathcal{L}_{\rm Diff}
=
\mathbb{E}_{\boldsymbol{y},\,\boldsymbol{\eta},\,n}
\left[
\left\|
\boldsymbol{\eta}
-
G_{\boldsymbol{w}_{\rm den}}(\boldsymbol{h}_{n},n,\boldsymbol{X})
\right\|_2^2
\right]
\end{equation}
where $n$ is sampled uniformly from $\{1,\dots,N\}$. After training, scenarios are generated by starting from $\boldsymbol{h}_{N}\sim\mathcal{N}(\boldsymbol{0},\boldsymbol{I})$ and iteratively applying the learned reverse denoising transitions until $\boldsymbol{h}_{0}$ is obtained, yielding the scenario $\hat{\boldsymbol{y}}=G_{\boldsymbol{w}_{\rm den}}(\boldsymbol{h}_{N},\boldsymbol{X})$.

\textit{Summary:}
Although the above three generative models differ in architecture and training objective, they can be incorporated into the proposed framework in a unified manner. Specifically, given the conditional input $\boldsymbol{X}$ and a random source variable $\boldsymbol{\eta}$ (e.g., a latent variable in VAE, Gaussian noise in GAN, or the initial noisy state in diffusion models), the generative model with parameters $\boldsymbol{w}$ can be abstractly represented as
$
\hat{\boldsymbol{y}} = G_{\boldsymbol{w}}(\boldsymbol{\eta}, \boldsymbol{X}).
$
By repeatedly sampling $\boldsymbol{\eta}$, a scenario set $
S(\boldsymbol{w})=\{\hat{\boldsymbol{y}}^{(s)}\}_{s=1}^{S}$
can be generated. These scenarios are subsequently passed to the scenario selection module for downstream dispatch optimization.

\textbf{Forward propagation of scenario selector:}
For decision-focused training, we first apply a random scenario selector to perform scenario selection. Specifically, given the generated scenario set $S(\boldsymbol{w})$, the selector randomly samples $K$ scenarios without replacement to form a reduced scenario set $S_{\rm r}(\boldsymbol{w})$, which is then fed into the downstream DUU module. Let $\mathcal{K}\subseteq \{1,\dots,S\}$ denote the set of selected scenario indices, with $|\mathcal{K}|=K$. The reduced scenario set can then be written as
$S_{\rm r}(\boldsymbol{w})=\{\hat{\boldsymbol{y}}^{(k)}\}_{k\in\mathcal{K}}$.

\textbf{Forward propagation of DUU module:}
To solve the two-stage DRO problem \eqref{eq:DRO_two_stage}, the main computational challenge lies in the worst-case expectation over the Wasserstein ambiguity set. For a fixed first-stage decision $\boldsymbol{x}$, the inner term $
\sup_{\mathbb{Q}\in\mathbb{D}} \mathbb{E}_{\boldsymbol{y}\sim\mathbb{Q}}
\left[Q(\boldsymbol{x},\boldsymbol{y})\right] $
admits the following strong dual reformulation \cite{skalyga2023distributionally}:
\begin{subequations}
\begin{align}
    &\inf_{\gamma\ge 0,\{\phi_k\}_{k=1}^{K}}
\left\{
\gamma\epsilon+\frac{1}{K}\sum_{k=1}^{K}\phi_k
\right\}\\
&\text{s.t.}\sup_{\boldsymbol{y}\in\Xi}
\left(
Q(\boldsymbol{x},\boldsymbol{y})
-\gamma \|\boldsymbol{y}-\boldsymbol{y}^{(k)}\|
\right)
\le \phi_k \quad \forall k \label{dro_dual_constraints}
\end{align}
\end{subequations}
where $\gamma$ and $\phi_k$ are auxiliary dual variables.
Moreover, since the second-stage problem \eqref{eq:DRO_two_stage_c}--\eqref{eq:DRO_two_stage_d} is a linear program (LP) in the variable $\boldsymbol{u}$ and the uncertainty $\boldsymbol{y}$ enters the constraints linearly, the recourse value function $Q(\boldsymbol{x},\boldsymbol{y})$ is convex in $\boldsymbol{y}$. Hence, the constraints \eqref{dro_dual_constraints} admit a tractable deterministic counterpart over the support set $\Xi$. In particular, when $\Xi$ is specified by componentwise lower and upper bounds and the Wasserstein metric is induced by the $1$-norm, the resulting reformulation is a linear program \cite{skalyga2023distributionally}.
The tractable counterpart of the two-stage Wasserstein-DRO dispatch is formulated as
\begin{subequations}\label{eq:dro_final_lp}
\begin{align}
\min_{\boldsymbol{x},\gamma\ge 0,\{\phi_k\}_{k=1}^{K}} \quad &
\mathcal{C}^{\rm DA}(\boldsymbol{x})
+\gamma\epsilon+\frac{1}{K}\sum_{k=1}^{K}\phi_k \\
\text{s.t.}\quad &
\boldsymbol{x}\ \text{satisfies}\ \eqref{eq:DA_genlim}\text{--}\eqref{eq:DA_linecap},\\
&
Q(\boldsymbol{x},\underline{\boldsymbol{y}})
-\gamma \|\boldsymbol{y}^{(k)}-\underline{\boldsymbol{y}}\|_1
\le \phi_k \quad \forall k\\
&
Q(\boldsymbol{x},\overline{\boldsymbol{y}})
-\gamma \|\boldsymbol{y}^{(k)}-\overline{\boldsymbol{y}}\|_1
\le \phi_k \quad \forall k\\
&
Q(\boldsymbol{x},\boldsymbol{y}^{(k)})\le \phi_k,\quad \forall k
\end{align}\label{lp_dro}
\end{subequations}
Importantly, the reformulated DUU problem in \eqref{eq:dro_final_lp} is explicitly parameterized by the generated scenario set
$S_{\rm r}(\boldsymbol{w})$.
This structure not only ensures tractability in the forward propagation but also enables the gradient backpropagation through the DUU module.

\textbf{Backward propagation of DUU module:}
We compute the gradient of the optimal dispatch solution and the resulting cost with respect to the scenario set $S_{\rm r}(\boldsymbol{w})$ using OptNet \cite{amos2017optnet}.
%, so that the operational feedback can be propagated to the upstream scenario generator.
Let the decision variable of \eqref{eq:dro_final_lp} be denoted by
$
\bar{\boldsymbol{x}}=
\left[
\boldsymbol{x},\gamma,\{\phi_k\}_{k=1}^{K}
\right]$
% and $S(\boldsymbol{w})=\{\boldsymbol{y}^{(s)}\}_{s=1}^{S}$ denote the scenario set generated by the scenario generator parameterized by $\boldsymbol{w}$. 
Then \eqref{eq:dro_final_lp} can be written in the following abstract LP form:
\begin{subequations}\label{eq:abstract_lp}
\begin{align}
\min_{\bar{\boldsymbol{x}}}\quad & \mathcal{C}^{\rm DA}(\bar{\boldsymbol{x}},S_{\rm r}(\boldsymbol{w}))\\
\textbf{s.t.}\quad &
\mathcal{F}^{\text{eq}}(\bar{\boldsymbol{x}},S_{\rm r}(\boldsymbol{w}))=\boldsymbol{0}\\
&
\mathcal{F}^{\text{iq}}(\bar{\boldsymbol{x}},S_{\rm r}(\boldsymbol{w}))\le \boldsymbol{0}
\end{align}
\end{subequations}
The Lagrangian function of \eqref{eq:abstract_lp} is
\begin{equation}
\begin{aligned}
    \mathcal{L}
=&
\mathcal{C}^{\rm DA}(\bar{\boldsymbol{x}},S_{\rm r}(\boldsymbol{w}))
+
(\boldsymbol{\lambda}^{\text{eq}})^{\top}
\mathcal{F}^{\text{eq}}(\bar{\boldsymbol{x}},S_{\rm r}(\boldsymbol{w}))
\\&+
(\boldsymbol{\lambda}^{\text{iq}})^{\top}
\mathcal{F}^{\text{iq}}(\bar{\boldsymbol{x}},S_{\rm r}(\boldsymbol{w}))
\end{aligned}
\end{equation}
where $\boldsymbol{\lambda}^{\text{eq}}$ and $\boldsymbol{\lambda}^{\text{iq}}$ are the dual variables associated with the equality and inequality constraints, respectively.

Since the optimal solution of \eqref{eq:abstract_lp} is generally not available in closed form, it cannot be directly differentiated with respect to the generated scenario set $S_{\rm r}(\boldsymbol{w})$. Instead, we characterize the dependence of the optimizer on $S(\boldsymbol{w})$ through the KKT conditions. Let
$
\tilde{\boldsymbol{x}}
=
\left[
\bar{\boldsymbol{x}},
\boldsymbol{\lambda}^{\text{eq}},
\boldsymbol{\lambda}^{\text{iq}}
\right]
$
collect the primal and dual variables. Then the KKT system can be compactly written as
\begin{equation}
\mathcal{K}(\tilde{\boldsymbol{x}},S(\boldsymbol{w}))
=
\left[
\begin{array}{c}
\nabla_{\bar{\boldsymbol{x}}}\mathcal{L} \\
\boldsymbol{\lambda}^{\text{iq}} \odot
\mathcal{F}^{\text{iq}}(\bar{\boldsymbol{x}},S_{\rm r}(\boldsymbol{w})) \\
\mathcal{F}^{\text{eq}}(\bar{\boldsymbol{x}},S_{\rm r}(\boldsymbol{w}))
\end{array}
\right]
=
\boldsymbol{0}
\label{eq:kkt_implicit}
\end{equation}
where $\odot$ denotes the element-wise product.

Assuming that the KKT point is locally unique and that the Jacobian
$
\partial \mathcal{K}/\partial \tilde{\boldsymbol{x}}
$
is nonsingular at the optimum, the implicit function theorem yields
\begin{equation}
\frac{\partial \tilde{\boldsymbol{x}}^{*}}{\partial S_{\rm r}(\boldsymbol{w})}
=
-\left(
\frac{\partial \mathcal{K}}
{\partial \tilde{\boldsymbol{x}}^{*}}
\right)^{-1}
\frac{\partial \mathcal{K}}
{\partial S_{\rm r}(\boldsymbol{w})}
\label{eq:implicit_grad_duu}
\end{equation}
This gives the sensitivity of the optimal primal-dual solution, and in particular the sensitivity of the day-ahead decision variable $\bar{\boldsymbol{x}}^{*}$ with respect to the generated scenario set $S_{\rm r}(\boldsymbol{w})$.

Accordingly, the gradient of the optimal day-ahead operational cost with respect to the generated scenario set can be obtained via the chain rule:
\begin{equation}
\frac{\partial \mathcal{C}^{\rm DA}(\bar{\boldsymbol{x}}^{*})}{\partial S_{\rm r}(\boldsymbol{w})}
=
\frac{\partial \mathcal{C}^{\rm DA}(\bar{\boldsymbol{x}}^{*})}{\partial \bar{\boldsymbol{x}}^{*}}
\frac{\partial \bar{\boldsymbol{x}}^{*}}{\partial S_{\rm r}(\boldsymbol{w})}
\label{eq:scenario_grad_da}
\end{equation}

Although the previous DA optimal solution has considered the worst-case expected RT cost, the decision-focused learning should also consider the quality of the ``actual'' RT decision under the realised scenario, so we need to introduce another OptNet layer to the backward process. In this layer, the day-ahead decision $\bar{\boldsymbol{x}}^{*}$ serves as the input parameter, and the optimal real-time decision $\boldsymbol{u}^{*}$ is obtained by solving the real-time dispatch problem. The resulting overall operational cost is
\begin{equation}
C^{*}
=
\mathcal{C}^{\rm DA}(\bar{\boldsymbol{x}}^{*})
+
\mathcal{C}^{\rm RT}(\boldsymbol{u}^{*}(\bar{\boldsymbol{x}}^{*}),\boldsymbol{y})
\end{equation}

For a fixed realization $\boldsymbol{y}$, once $\bar{\boldsymbol{x}}^{*}$ is given, the optimal real-time decision $\boldsymbol{u}^{*}$ is determined by the real-time optimization problem and can therefore be regarded as an implicit function of $\bar{\boldsymbol{x}}^{*}$. Applying OptNet to the real-time dispatch layer yields the gradient $
{\partial \boldsymbol{u}^{*}}/{\partial \bar{\boldsymbol{x}}^{*}}$.
Accordingly, the gradient contribution of the RT layer can be computed by the chain rule as
\begin{equation}
\frac{\partial C^{*}}{\partial \bar{\boldsymbol{x}}^{*}}
=
\frac{\partial \mathcal{C}^{\rm DA}}{\partial \bar{\boldsymbol{x}}^{*}}
+
\frac{\partial \mathcal{C}^{\rm RT}}{\partial \boldsymbol{u}^{*}}
\frac{\partial \boldsymbol{u}^{*}}{\partial \bar{\boldsymbol{x}}^{*}}
\label{eq:grad_total_x}
\end{equation}

Combining \eqref{eq:grad_total_x} with \eqref{eq:implicit_grad_duu}, the gradient of the overall operational cost with respect to the generated scenario set is
\begin{equation}
\frac{\partial C^{*}}{\partial S_{\rm r}(\boldsymbol{w})}
=
\frac{\partial C^{*}}{\partial \bar{\boldsymbol{x}}^{*}}
\frac{\partial \bar{\boldsymbol{x}}^{*}}{\partial S_{\rm r}(\boldsymbol{w})}
\label{eq:grad_total_s}
\end{equation}
Finally, these gradients are propagated backward through the scenario selector and generator to update its parameters $\boldsymbol{w}$.

\textbf{Backward propagation of scenario selector:}
Although the random index sampling itself is non-differentiable, it does not block gradient backpropagation to the upstream generative model. This is because the reduced scenario set $S_{\rm r}(\boldsymbol{w})$ is constructed from $S(\boldsymbol{w})$ through a standard indexing operation. Therefore, during backpropagation, the gradient of the downstream operational cost with respect to the generated scenarios satisfies
\begin{equation}
\frac{\partial C^*}{\partial \hat{\boldsymbol{y}}^{(s)}}=
\begin{cases}
\frac{\partial C^*}{\partial \hat{\boldsymbol{y}}^{(s)}_{\rm r}}, & s\in\mathcal{K}\\
0, & s\notin\mathcal{K}
\end{cases}
\end{equation}
Hence, the random scenario selector preserves a valid gradient path from the downstream optimization layer to the selected generated scenarios, enabling decision-focused training of the upstream generator.

\textbf{Backward propagation of generative model:}
After obtaining the gradient of the overall operational cost with respect to $S_{\rm r}(\boldsymbol{w})$, the remaining task is to backpropagate this signal through the generative model and update its parameters $\boldsymbol{w}$. Although the three considered generative models share the same forward interface through $S(\boldsymbol{w})$, their backward updates differ according to their internal generation mechanisms.

\textit{1) VAE:}
For the VAE, scenario generation only involves the decoder. Therefore, during decision-focused fine-tuning, the downstream gradient is backpropagated through the decoder to update $\boldsymbol{w}_{\rm dec}$.

\textit{2) GAN:}
% For the GAN, only the generator participates in scenario generation, while the discriminator is used only in adversarial training. Therefore, the operational cost gradient is backpropagated only through the generator $G_{\boldsymbol{w}_G}$, and only the generator parameters $\boldsymbol{w}_G$ are updated.
For the GAN, the generator and discriminator are first pretrained using the standard adversarial loss to learn the data distribution. After pretraining, the discriminator is no longer used in the decision-focused fine-tuning stage, since only the generator participates in scenario generation. Therefore, the downstream operational cost gradient is backpropagated only through the generator $G_{\boldsymbol{w}_G}$, and only the generator parameters $\boldsymbol{w}_G$ are updated during decision-focused fine-tuning.

\textit{3) Diffusion model:}
For the diffusion model, backward propagation is more involved. In standard DDPM training, the denoising network is optimized by sampling a diffusion step and minimizing the corresponding noise prediction error. In contrast, under the proposed decision-focused framework, the scenario sent to the downstream decision-making module is generated through the full reverse denoising process over $N$ steps. In principle, the operational cost gradient should therefore be backpropagated through all reverse denoising steps.
However, preserving the entire reverse trajectory leads to high memory consumption and severe gradient attenuation. To address this issue, we adopt a truncated backpropagation strategy. Specifically, instead of retaining all $N$ denoising steps, we preserve only the last $n_{\rm tru}$ steps in the computational graph, where $n_{\rm tru} \ll N$. Since the outputs of the final denoising steps are already close to the generated scenario, these steps have the most direct influence on the downstream decision. Consequently, the decision-focused gradient is backpropagated only through these final $n_{\rm tru}$ steps to update $\boldsymbol{w}_{\rm den}$.

\textit{Summary:}
Once the gradient $\partial C^*/\partial S_{\rm r}(\boldsymbol{w})$ is obtained from the downstream optimization layers, it is propagated back through the generative model according to its architecture: only the decoder are updated for the VAE, only the generator is updated for the GAN, and truncated backpropagation through the final denoising steps is used for the diffusion model. This completes the decision-focused training of the scenario generation module.

\subsection{Decision-focused Scenario Selection}\label{sec:scenario_reduction}

\begin{figure}
    \centering
    \includegraphics[width=1\linewidth]{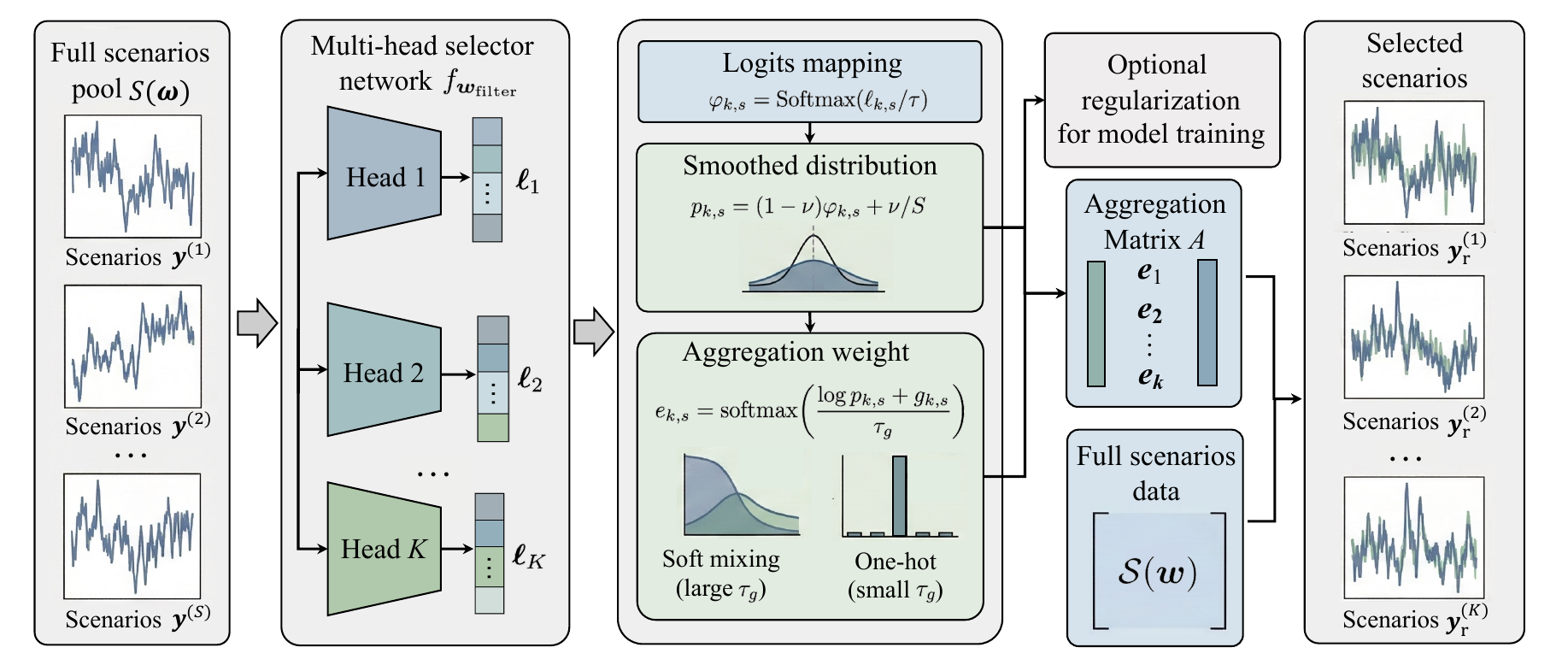}
    \caption{The framework of the proposed scenario selector}
    \label{fig:selector}
\end{figure}

The size of 
$\mathcal{S}(\boldsymbol{w})$ directly affects the computational tractability of decision-focused training, making scenario selection essential. When a subset 
$\mathcal{S}_{\rm r}\subseteq \mathcal{S}$ with 
$|\mathcal{S}_{\rm r}|=K$ is selected, the empirical distribution induced by the retained scenarios generally deviates from that of the original full scenario set.
As a consequence, the corresponding Wasserstein ambiguity set constructed from $\mathcal{S}_{\rm r}$ also differs from that induced by $\mathcal{S}$. In general, solving the DRO problem based on $\mathcal{S}_{\rm r}$ is expected to yield inferior performance compared with using $\mathcal{S}$, due to the loss of information during scenario selection. To this end, we further introduce a data-driven scenario selector parameterized by $\boldsymbol{\theta}$, as shown in Fig. \ref{fig:selector}, which is trained in a decision-focused manner.

The selector outputs $K$ selection distributions over the original pool, one for each reduced scenario index $k\in\{1,\dots,K\}$.
The selector network maps the scenario pool to a matrix of logits $\boldsymbol{\ell}=f_{\boldsymbol{w}_{\rm selector}}(\mathcal{S})\in\mathbb{R}^{K\times S}$, where $\ell_{k,s}$ is the logit score of selecting the original scenario $\boldsymbol{y}^{(s)}$ for the $k$-th reduced scenario. A temperature-scaled softmax yields a probability vector over the $S$ candidates for each $k$.
To encourage exploration and mitigate early collapse, we apply uniform smoothing with coefficient $\nu$:
\begin{equation}
p_{k,s}=(1-\nu)\varphi_{k,s}+\nu/S,\qquad \sum_{s=1}^{S}p_{k,s}=1
\end{equation}
\textbf{Training of scenario selector:} To enable decision-focused learning with gradient-based methods, we employ a Gumbel--Softmax reparameterization to obtain a continuous weight vector $\boldsymbol{e}_k$ for each head $k$:
\begin{subequations}
\begin{alignat}{2}
&e_{k,s}=\mathrm{softmax}\!\left(\frac{\log p_{k,s}+g_{k,s}}{\tau_g}\right)\\
&g_{k,s}=-\log(-\log u_{k,s}),\ u_{k,s}\sim{\rm Uniform}(0,1)
\end{alignat}
\end{subequations}
This reparameterization provides a differentiable surrogate for discrete sampling or Top-K selection from $\{p_{k,s}\}_{s=1}^S$, both of which are non-differentiable and therefore incompatible with backpropagation through the downstream optimization loss.
Denote the stacking of $\{\boldsymbol{e}^\top\}_{k=1}^{K}$ as
$\boldsymbol{E}=[\boldsymbol{e}_1^\top,\dots,\boldsymbol{e}_{K}^\top]\in\mathbb{R}^{K\times S}$. The reduced scenarios are constructed as convex combinations of the base scenarios pool:
\begin{equation}
\boldsymbol{y}_{\rm r}^{(k)}=\sum_{s=1}^{S}E_{k,s}\boldsymbol{y}^{(s)},\qquad k\in\{1,\dots,K\}
\end{equation}
yielding $\mathcal{S}_{\rm r}=\{\boldsymbol{y}_{\rm r}^{(k)}\}_{k=1}^{K}$. As $\tau_g\to 0$, each $\boldsymbol{e}_k$ approaches a one-hot vector, and the construction recovers discrete scenario selection while remaining differentiable during training. 

To discourage degenerate solutions in which multiple heads collapse to the same scenarios, we augment the downstream loss with a diversity regularizer. The training objective is $\mathcal{L}=C^{*}+\lambda\,\mathcal{L}_{\rm reg}$, where $\boldsymbol{p}_k=[p_{k,1},\dots,p_{k,S}]^\top$ is the selection distribution of head $k$ and $\lambda\ge 0$. We consider the following three kinds of regularization items in our work: inner production, entropy, and KL divergence.
% \begin{align}
% \mathcal{L}_{\rm reg}^{\rm Ip} &= \sum_{k<k'} \boldsymbol{p}_k^\top \boldsymbol{p}_{k'}\\
% \mathcal{L}_{\rm reg}^{\rm KL} &= -\sum_{k<k'} \Big(\mathrm{KL}(\boldsymbol{p}_k\|\boldsymbol{p}_{k'})+\mathrm{KL}(\boldsymbol{p}_{k'}\|\boldsymbol{p}_k)\Big) \\
% \mathcal{L}_{\rm reg}^{\rm Ent} &= -\sum_{k=1}^{K} H(\boldsymbol{p}_k)
% \end{align}
% where $\mathrm{KL}(\boldsymbol{p}\|\boldsymbol{q})=\sum_{s=1}^{S} p_s\log\!\left(\frac{p_s}{q_s}\right)$ and $H(\boldsymbol{p})=-\sum_{s=1}^{S} p_s\log p_s$.

% \textbf{Evaluation of scenario selector} The reduced set can be formed either by using the learned weights $\boldsymbol{E}$ to produce convex combinations (continuous reduction) or by selecting the highest-probability scenarios implied by $\{p_{k,s}\}$ (discrete reduction). 
% This design enables decision-focused training while yielding a practically usable reduced scenario set for downstream DUU. 

\textbf{Evaluation of scenario selector} At evaluation time, we adopt discrete scenario selection according to ${p_{k,s}}$, so that the retained scenarios are actual samples generated by the generative model. This avoids introducing artificial interpolated scenarios and ensures that the reduced scenario set can be directly used in downstream DUU.

\section{Case Studies}
\label{case}

\subsection{Experimental Setups}
\subsubsection{Dataset}
This study evaluates the proposed framework on the IEEE 14-bus system, using load data collected from southern China \cite{li2025large}. These load data are collected in different zones within a city and show strong spatial correlations. Following the practices as \cite{feng2024out} and \cite{park2025load}, the load data and the system configurations are proportionally scaled to ensure operational feasibility.
The dataset covers the period from 2022 to 2023, with data from 2022 used as the training and validation datasets, and data from 2023 reserved for evaluation. The training and validation datasets are randomly split in an 80\% to 20\% proportion.

\subsubsection{Baseline}
As for the forecasting model, we evaluate eight models, including both traditional parametric and non-parametric forecasting approaches. As previously mentioned, it is challenging for these two types of models to jointly forecast the joint distribution of loads from different buses, therefore, we conduct separate forecasts for each. For the generative models, we test GAN, VAE, and Diffusion models in both separate and joint forecasting settings simultaneously.
For the IEEE 14-bus system, which has 11 load buses, we train 11 individual forecasting models for the separate setting and one model for the joint forecasting setting.

In the context of decision-making, we consider several settings. The first is solving an optimization problem with perfect forecasts. It is an idealized scenario that is impossible in practice and serves as the performance upper bound. As there is no uncertainty under the deterministic forecast, this setting simply dispatches based on fixed up and down reserve capacity, typically using a predetermined percentage, as described in \cite{mathieu2014quantitative}. 
In addition, we will apply the scenarios generated by eight different forecasting models to DRO, which is represented by \eqref{eq:DRO_two_stage_d}, and select the radii of the ambiguity set for these scenarios using the validation dataset. We compare the following methods: accuracy-oriented scenario generation with random scenario selection (AO), decision-focused scenario generation with random scenario selection (DF), and decision-focused scenario generation with the proposed selectors trained with different regularization terms, namely DF (Selector-KL), DF (Selector-Inner), and DF (Selector-Entropy).

For the comparison of scenario selection methods, we will train the selectors proposed in this work using three different regularization techniques and compare their performance. Specifically, we compare our proposed scenario selection approach with three widely adopted methods: hierarchical clustering \cite{liu2017hierarchical}, K-means \cite{lin2021stochastic}, and K-medoids \cite{pei2022precise}, as well as random selection.

\subsubsection{Evaluation Metrics}
This work evaluates forecasting accuracy from both deterministic and probabilistic aspects. For the deterministic aspect, forecasts are assessed using the mean absolute error (MAE) and root mean square error (RMSE). For the probabilistic aspect, forecast quantiles from 0.1 to 0.9 with an interval of 0.1 are evaluated using the pinball loss. % as follows:
Additional details, including the source code and relevant datasets, are available in our GitHub repository \href{https://github.com/hkuedl/Cost-oriented-Generative-Model}{https://github.com/hkuedl/Cost-oriented-Generative-Model}.
%Ingore angles and flow with a copper plate model Copper Plate problem \cite{viens2025optimal}

\subsection{Forecasting performance}

\begin{figure}
    \centering
    \includegraphics[width=1\linewidth]{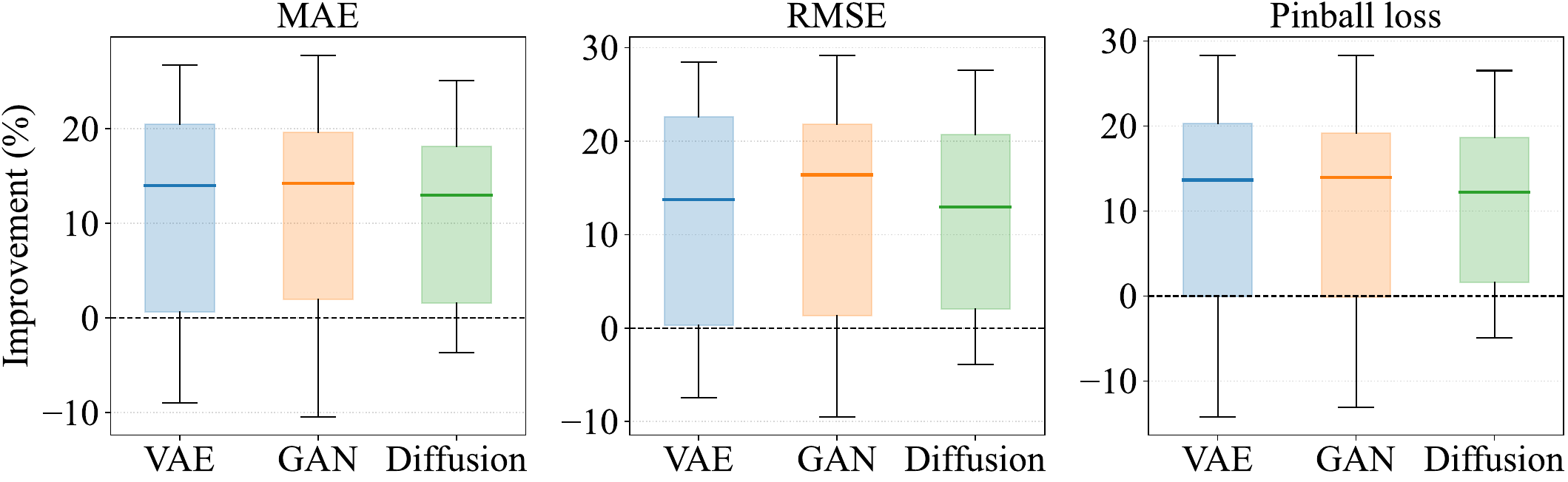}
    \caption{Accuracy comparison between the generative approach and parametric approach}
    \label{fig:box_comparison_parametric}
\end{figure}

\begin{figure}
    \centering
    \includegraphics[width=1\linewidth]{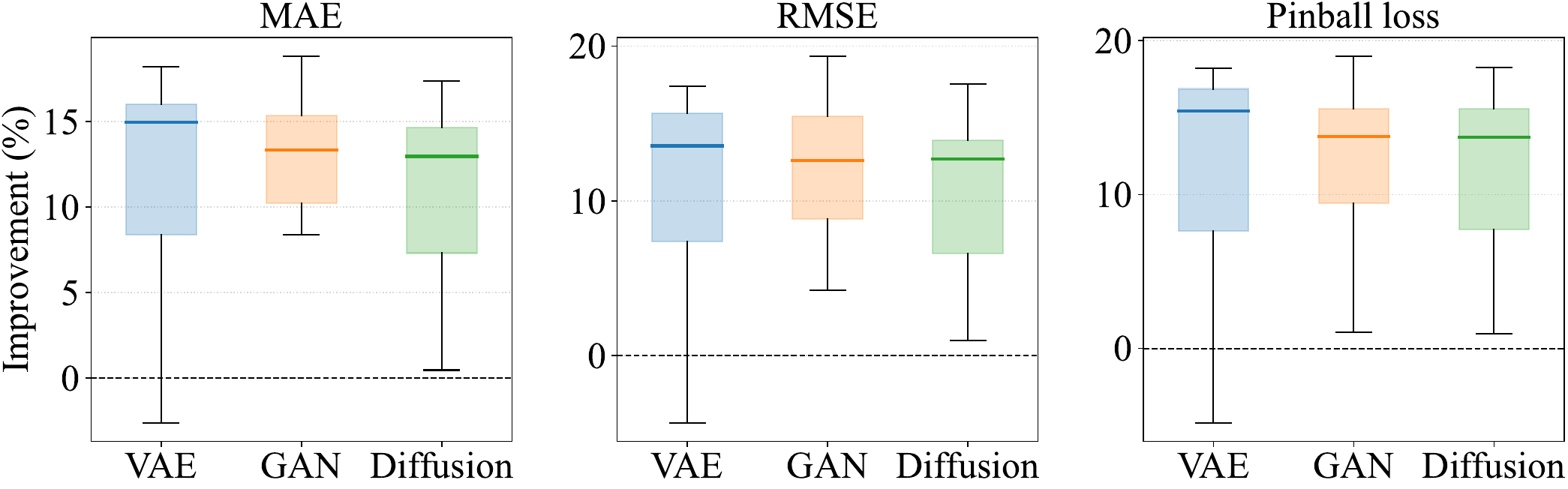}
    \caption{Accuracy comparison between the generative approach and non-parametric approach}
    \label{fig:box_comparison_non_parametric}
\end{figure}

\begin{figure}
    \centering
    \includegraphics[width=1\linewidth]{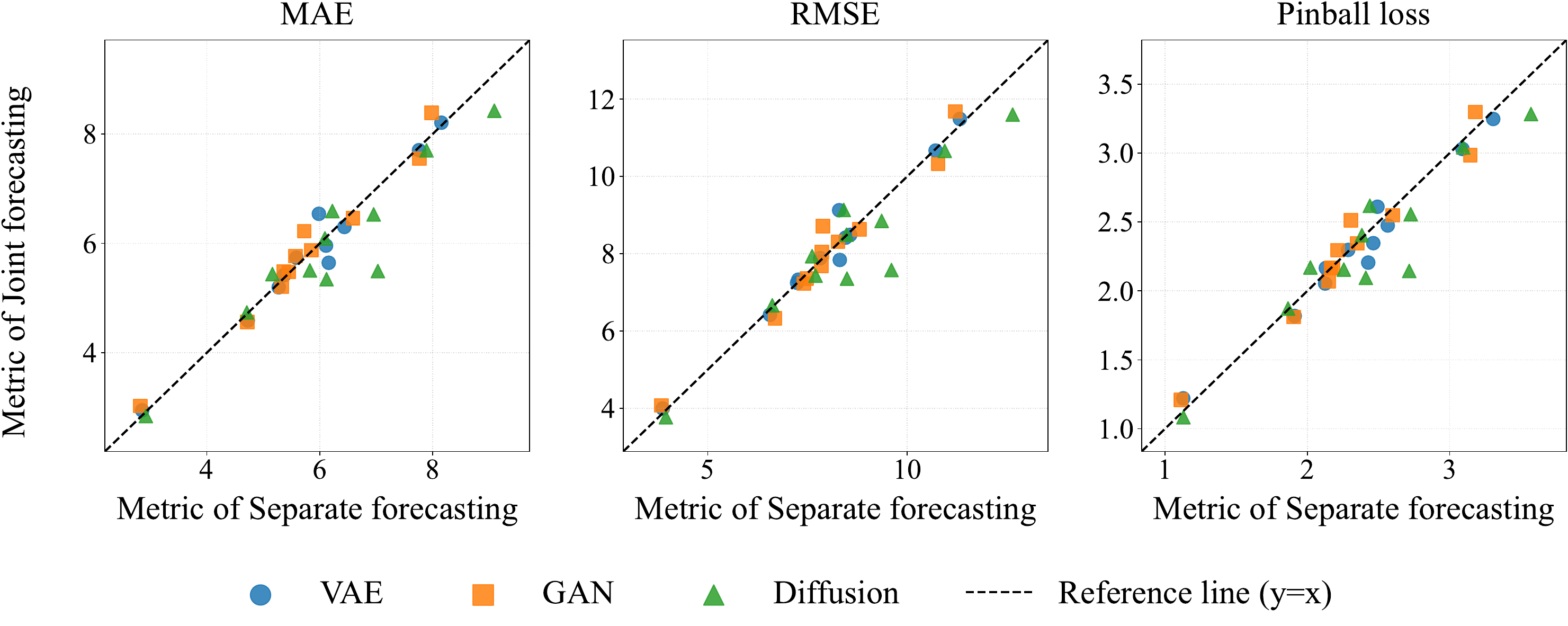}
    \caption{The accuracy comparison of separate forecasting and joint forecasting}
    \label{fig:y_x}
\end{figure}

Figs. \ref{fig:box_comparison_parametric} and \ref{fig:box_comparison_non_parametric} compare the accuracy improvement of VAE, GAN, and diffusion over the benchmark methods in terms of MAE, RMSE, and pinball loss. Overall, all three generative models achieve mostly positive improvements in both the parametric and non-parametric settings, indicating that the generative approach consistently outperforms the benchmarks. 
%Compared with the parametric approach, the improvements are generally larger but also more dispersed. In contrast, the gains over the non-parametric approach are more concentrated, suggesting more stable performance. The differences among VAE, GAN, and diffusion are relatively small, and no single model consistently dominates across all metrics.
Fig.~\ref{fig:y_x} compares separate forecasting and joint forecasting with generative models. The x-axis and y-axis denote the results of separate forecasting and joint forecasting, respectively, while different markers represent VAE, GAN, and diffusion models. The dashed diagonal line $y=x$ indicates identical performance. Points below the line mean joint forecasting achieves lower error. Overall, most points are close to the diagonal, with a slight tendency to lie below it. This indicates that joint forecasting achieves marginal forecasting metrics comparable to, and in some cases slightly better than, those of separate forecasting.

% Overall, most points lie lower than the diagonal, suggesting that the joint forecasting approach has better forecasting performance, meaning that considering correlations is helpful for forecasting.

% \begin{figure}[t]
%     \centering

%     \includegraphics[width=0.98\linewidth]{pdf_figures/box_parametric.pdf}
%     \par\smallskip
%     \centerline{\footnotesize (a) Parametric approach}
%     \medskip

%     \includegraphics[width=0.98\linewidth]{pdf_figures/box_non_parametric.pdf}
%     \par\smallskip
%     \centerline{\footnotesize (b) Non-parametric approach}
%     \caption{Accuracy comparison between the generative approach and benchmark methods.}
%     % \caption{Accuracy comparison between the generative approach and benchmark methods: (a) comparison with the parametric approach and (b) comparison with the non-parametric approach.}
%     \label{fig:box_comparison}
% \end{figure}

\begin{figure}[t]
    \centering

    \includegraphics[width=0.8\linewidth]{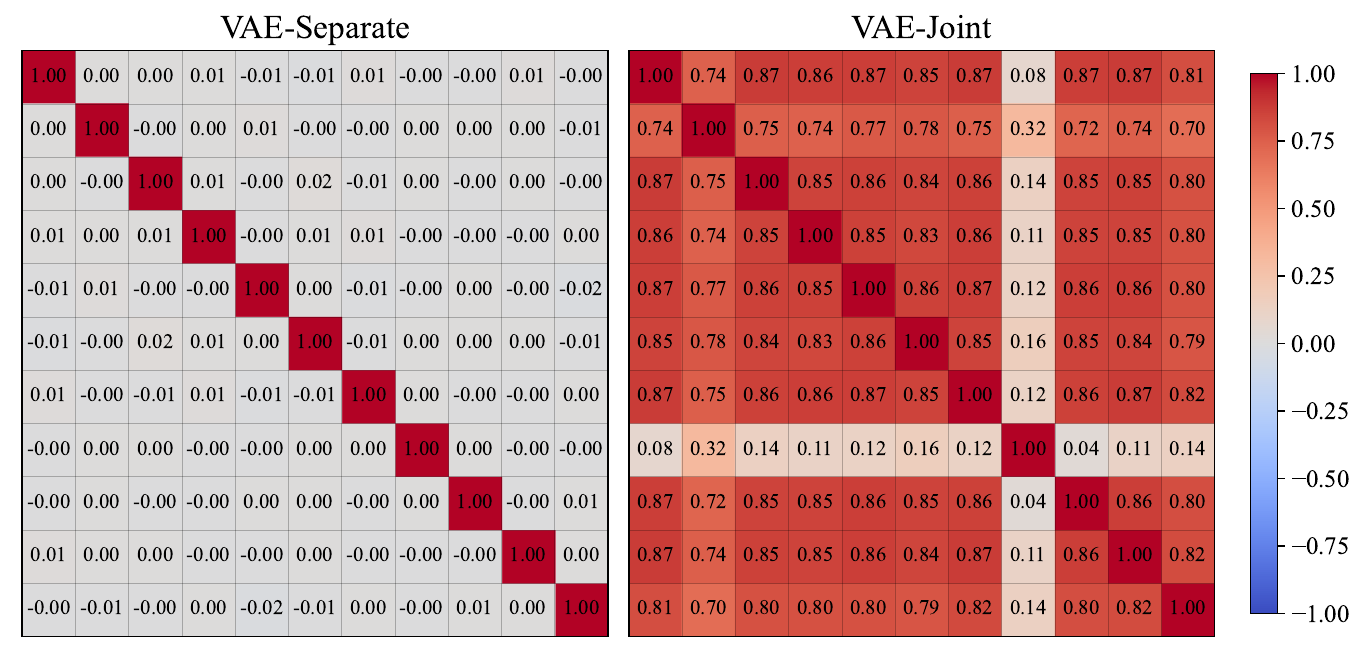}
    
    \vspace{0.2em}
    {\small (a) VAE}
    \vspace{0.8em}

    \includegraphics[width=0.8\linewidth]{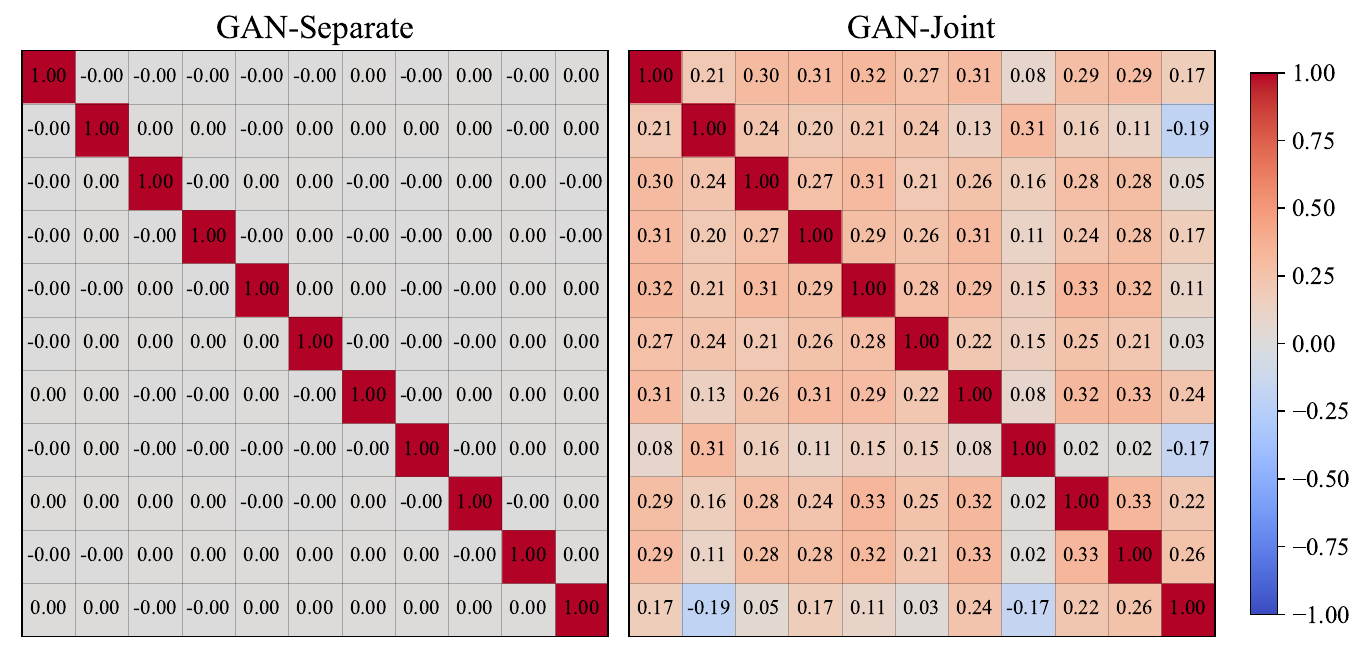}
    
    \vspace{0.2em}
    {\small (b) GAN}
    \vspace{0.8em}

    \includegraphics[width=0.8\linewidth]{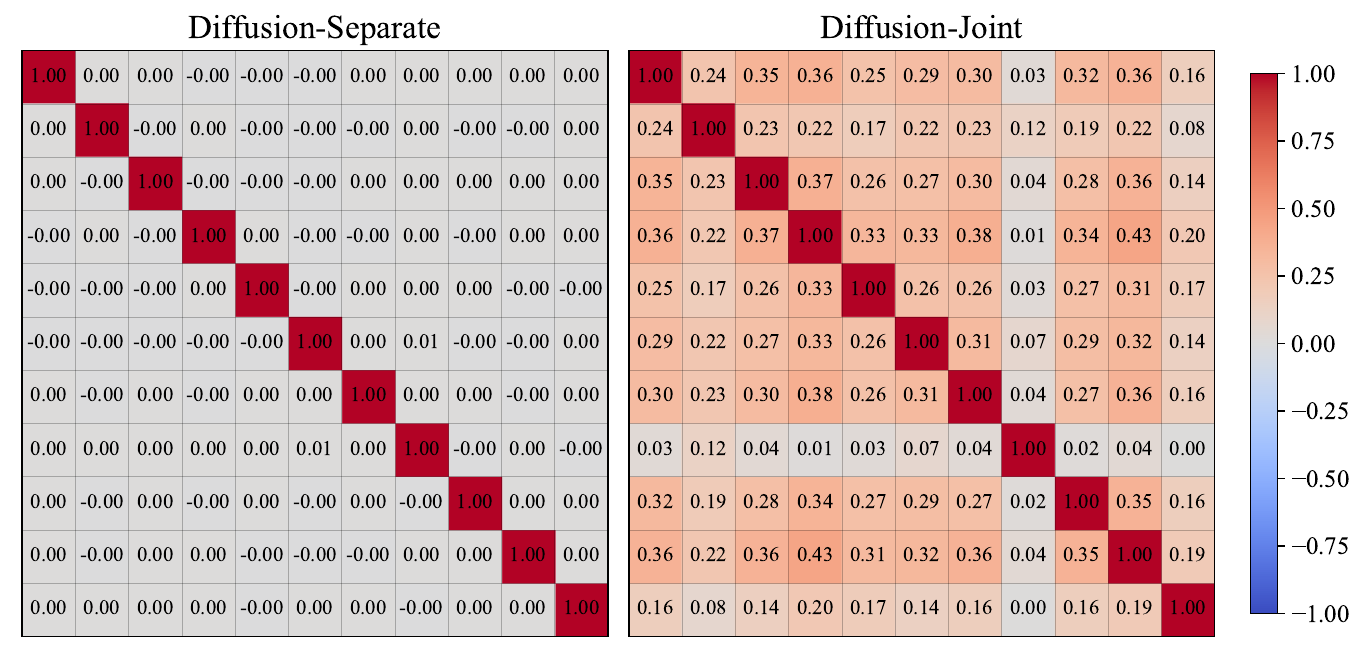}
    
    \vspace{0.2em}
    {\small (c) Diffusion}
    \caption{Correlation between different nodes learned by different generative models within separate and joint forecasting: (a) VAE (b) GAN, and (c) Diffusion.}
    \label{fig:heatmap_all}
\end{figure}

\begin{figure}[t]
    \centering

    \includegraphics[width=0.48\linewidth]{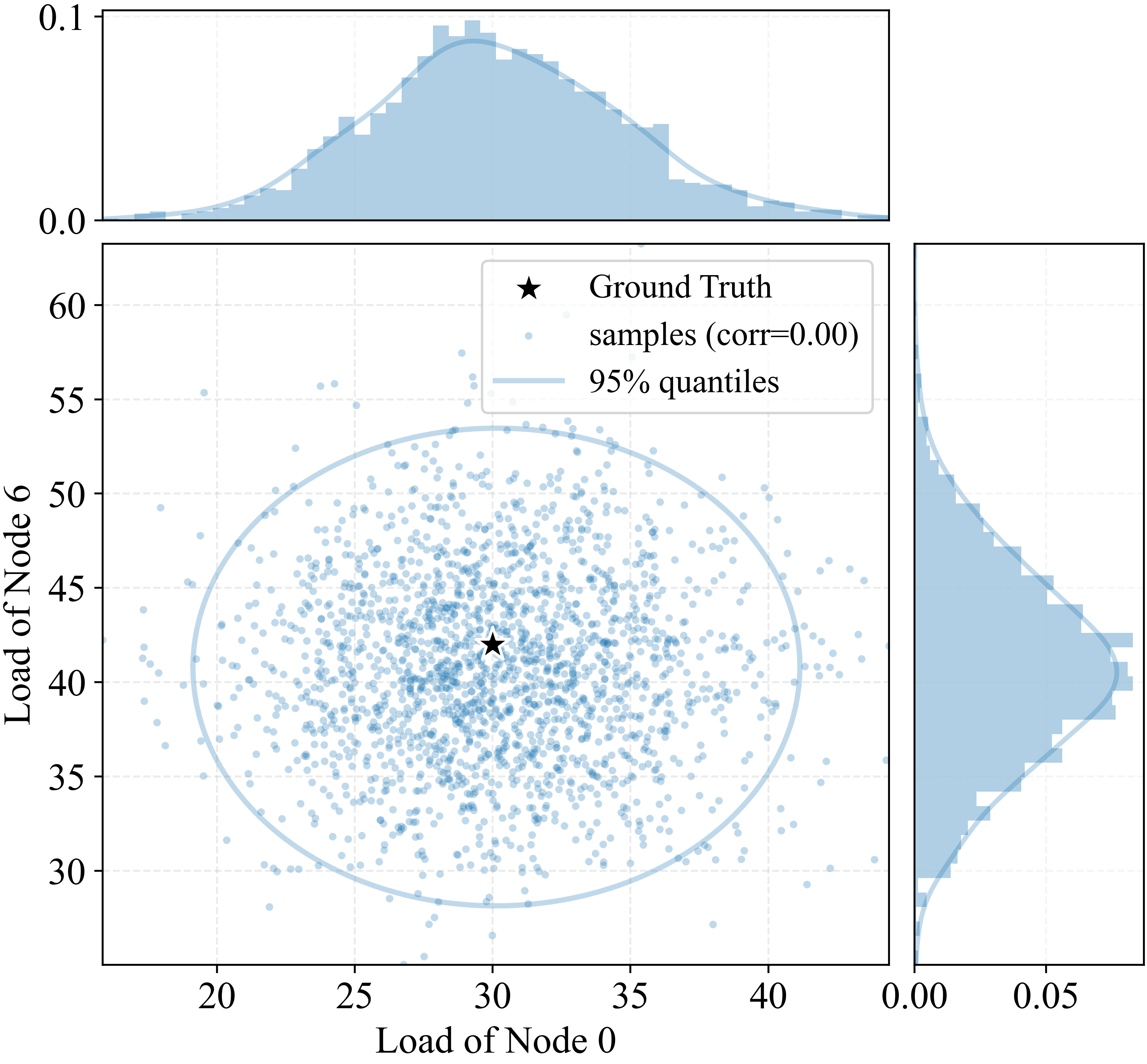}
    \hfill
    \includegraphics[width=0.48\linewidth]{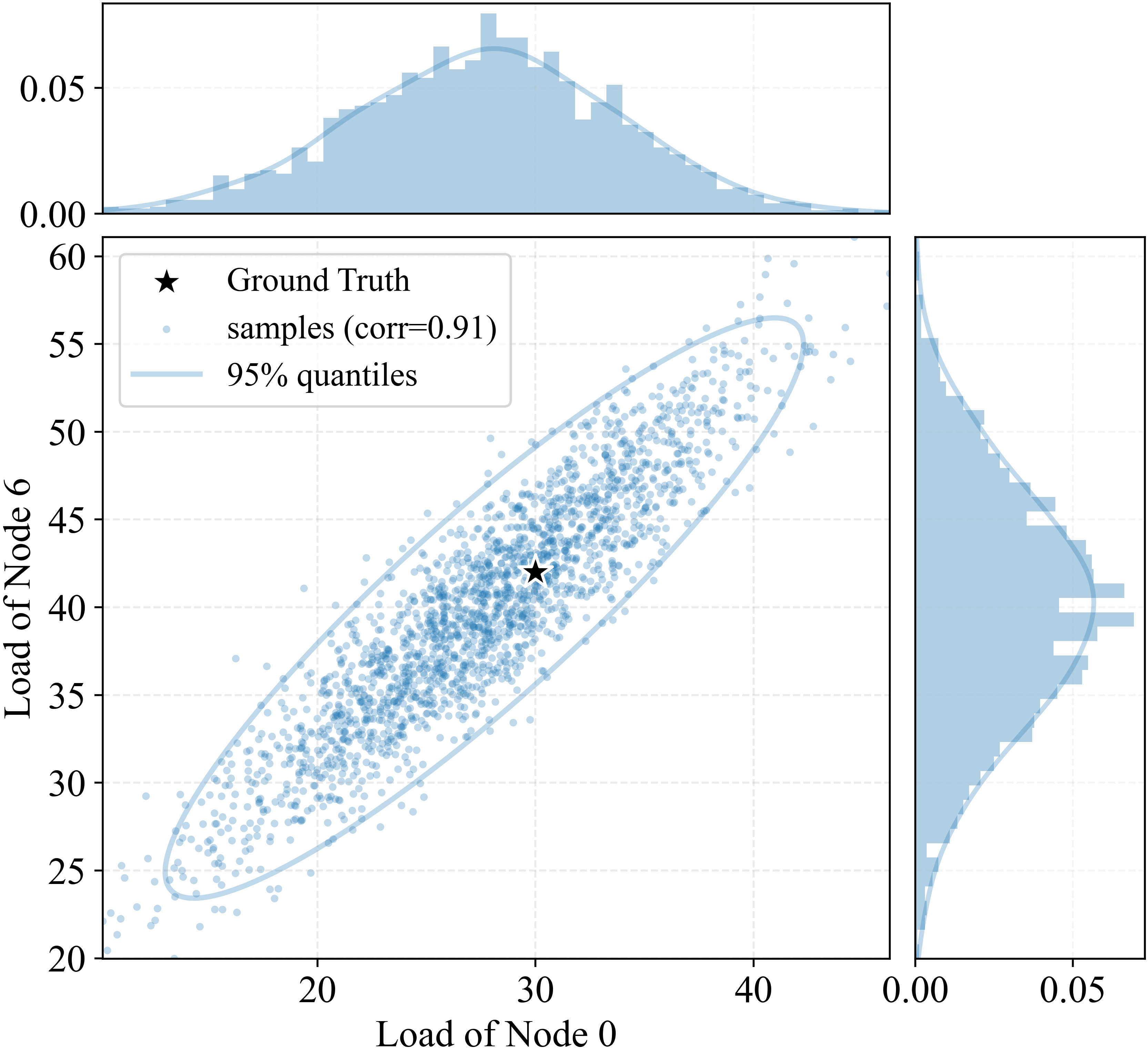}

    \par\smallskip
    \makebox[0.48\linewidth][c]{\small (a) Separate forecasting}
    \hfill
    \makebox[0.48\linewidth][c]{\small (b) Joint forecasting}

    \caption{Examples of marginal and joint distributions learned by different forecasting approaches: (a) separate forecasting and (b) joint forecasting.}
    \label{fig:marginals_comparison}
\end{figure}

% The correlation between different nodes learned by different generative models under separate and joint forecasting is illustrated in Fig. \ref{fig:heatmap_all}. We can observe that, under separate forecasting, the residuals of loads at different nodes are almost uncorrelated, indicating that separate forecasting cannot capture the dependency among nodes. In contrast, joint forecasting is able to learn relatively consistent correlation patterns.
% For VAE, GAN, and Diffusion, one node shows consistently weaker correlations with the other nodes.

Fig.~\ref{fig:heatmap_all} compares the cross-node correlation patterns of scenarios generated by separate and joint forecasting. Under separate forecasting, scenarios at different nodes are sampled independently, leading to weak off-diagonal correlations. In contrast, joint forecasting produces more evident cross-node dependence across the generated scenarios.
As shown in Fig.~\ref{fig:marginals_comparison}, the separately generated scenarios are more evenly scattered around the two marginal directions, leading to an approximately circular cloud. In contrast, the jointly generated scenarios show an elliptical shape, indicating that the two nodes tend to vary in a more coordinated way in the generated samples. Therefore, the joint approach reduces some extreme combinations caused by independent sampling, such as one node taking a very low value while the other takes a very high value.

% As shown in Fig.~\ref{fig:marginals_comparison}, the scenarios generated by the separate forecasting approach form an approximately circular pattern, while those generated by the joint forecasting approach exhibit an elliptical shape because the correlation between nodes is preserved. The separate forecasting approach also produces some unrealistic scenarios, such as one node having a very low load while the other has a very high load. By contrast, the joint forecasting approach yields a wider and more reasonable range of scenarios for each node and better captures the fact that loads at different nodes may increase or decrease simultaneously. 

%For example, for Node 0, the sampled range under joint forecasting is approximately 15-45 MW, compared with 20-40 MW under separate forecasting. This suggests that joint forecasting can better characterize extreme yet realistic scenarios.

\subsection{Operation Cost Performance}
\begin{figure}[t]
    \centering
    \includegraphics[width=0.95\linewidth]{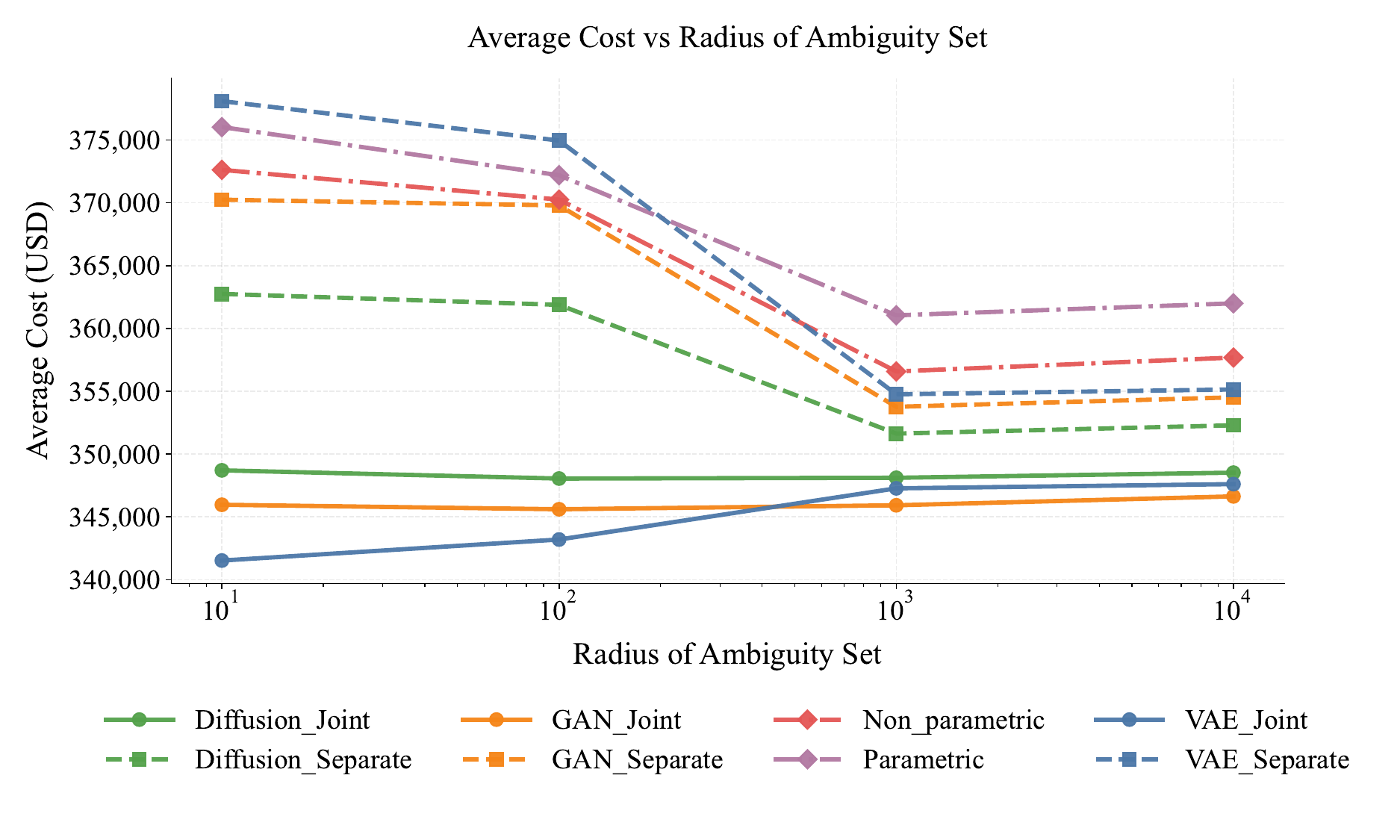}
    \caption{The average costs on the validation dataset for eight forecasting settings with different ambiguity set radii.}
     \label{case_fig9}
\end{figure}

The average costs on the validation dataset for eight forecasting settings with different ambiguity set radii are drawn in Fig. \ref{case_fig9}. It can be seen that, for all radii, the joint forecasting approaches that incorporate correlation information consistently achieve lower costs than the separate forecasting approaches. In addition, as the ambiguity set radius decreases, the costs of the separate forecasting approaches increase significantly. This is because the distributions learned by separate forecasting deviate more from the true distribution, and a small ambiguity set may fail to capture the true distribution. By contrast, the joint forecasting approaches remain relatively stable overall, although VAE-Joint shows a moderate increase in cost as the ambiguity set radius grows (still lower than the separate forecasting setting). On the other hand, when the radius becomes too large, the costs of both types of methods increase slightly, since the ambiguity set may then include unrealistic distributions, forcing the robust optimization model to adopt more conservative decisions.

\begin{figure}[t]
    \centering
    \includegraphics[width=0.98\linewidth]{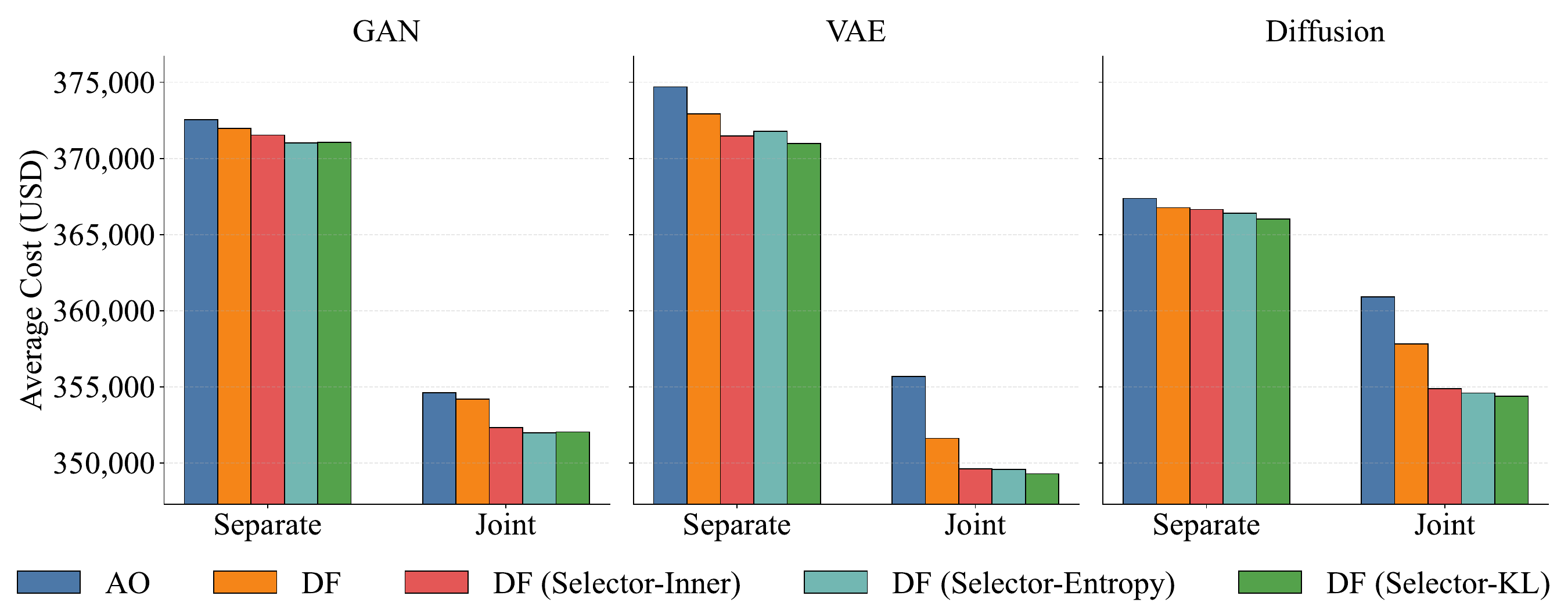}
    \caption{The average costs of AO, DF, and DF(Selector) methods for the generative models.}
     \label{case_fig10}
\end{figure}

The comparison of the average costs of AO, DF, and DF (Selector) methods for the generative models is shown in Fig. \ref{case_fig10}.
Across different generative models, the proposed decision-focused framework effectively reduces operational costs. Moreover, the selection is more pronounced in the joint forecasting setting. This is because scenarios generated under the joint forecasting setting tend to be more concentrated and require fewer model updates.
%In contrast, separate forecasting models necessitate simultaneous updates of generative models for different nodes, increasing training complexity. 
For various regularization items and different generative models, our selector can further reduce costs beyond what is achieved by decision-focused scenario generation alone.

The cost comparison between different scenario selection methods of eight forecasting settings is listed in Table \ref{table:cosr_comparison}. It is noted that all these scenario selection methods are applied to the generative model trained in a decision-focused approach.
We observe that existing scenario selection methods exhibit unstable performance, sometimes even performing worse than random selection. In contrast, our model, which can be trained using a decision-focused approach, achieves better operational cost performance across different regularization items.

\subsection{Efficiency analysis}
\begin{figure}[t]
    \centering
    \includegraphics[width=1\linewidth]{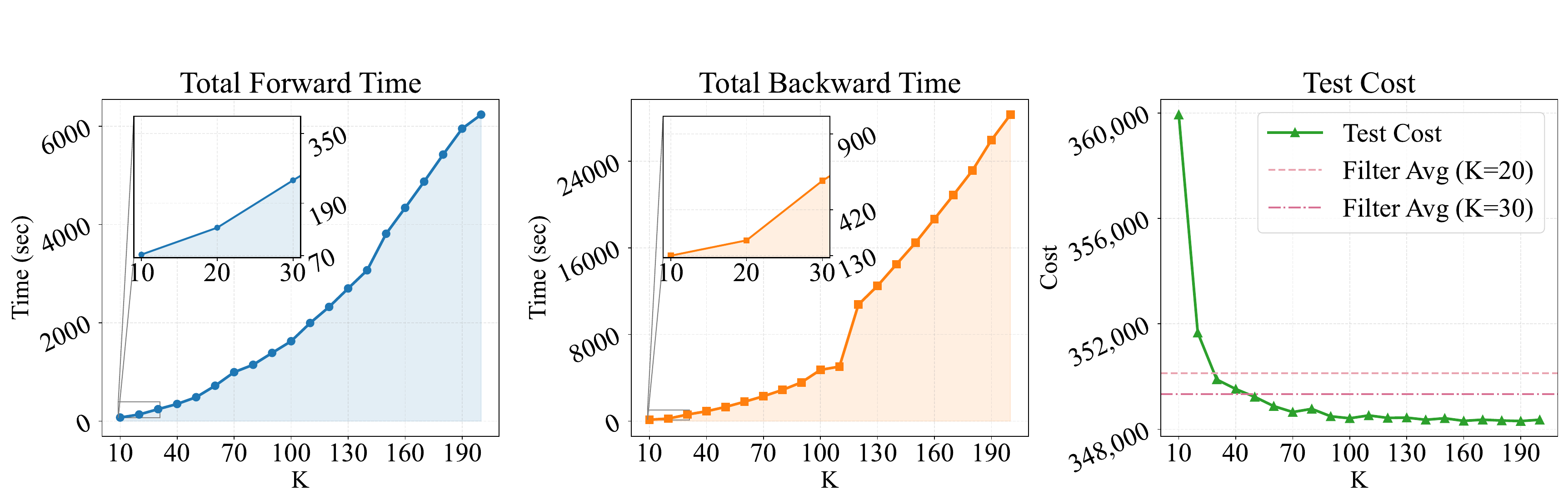}
    \caption{The total forward and backward time, and costs for various $K$ within a simplified system.}
     \label{case_fig11}
\end{figure}

%Due to the memory limitation of our computing devices (32 GB), it is challenging to analyze the computation time and operational cost without scenario reduction when $K$ is large.
Due to the excessive computational complexity to implement all case studies, we report the total forward and backward propagation time, together with the resulting test cost, for the simplified system in Fig.~\ref{case_fig11}.
%Fig.~\ref{case_fig11} presents the total forward and backward propagation time, together with the resulting test cost, for the simplified system. 
In this setting, the power flows among different buses are ignored, and only the overall energy balance is considered as a copper plate model \cite{viens2025optimal}. 
The results show that the proposed method can reduce computation time by 86.67\% with only a 0.26\% increase in cost.

% When the number of samples changes from K=100 to K=30, our method can achieve an 86.67\% time reduction with only a 0.26% increase in cost.

% both forward and backward propagation time with only a small number of samples, while achieving a test cost comparable to that obtained with a much larger sample set.

\begin{figure}[t]
    \centering
    \includegraphics[width=0.98\linewidth]{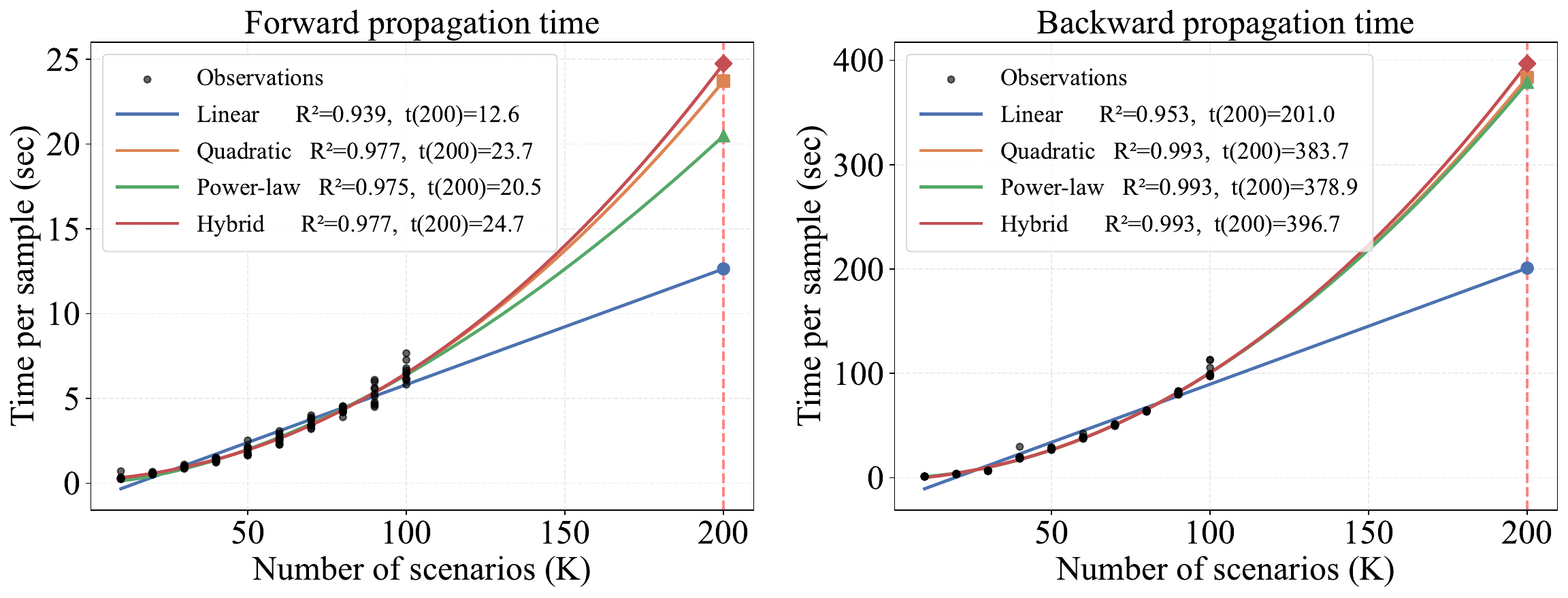}
    \caption{The forward and backward time per sample for VAE in the joint forecasting setting.}
     \label{case_fig12}
\end{figure}

Fig.~\ref{case_fig12} shows the forward and backward propagation time per sample of the VAE in the joint forecasting setting as the number of scenarios $K$ increases. We fit the observed runtimes using four models: linear, quadratic, power-law, and hybrid. For each $K$, 10 runs are conducted to improve robustness. The extrapolation results indicate a substantial computational burden at large $K$. Under the hybrid fit, when $K=200$, the forward and backward propagation time per sample are estimated to be 24.7 and 396.7 seconds, respectively. The reason we only tested up to K = 100 is that when K becomes larger, it may exceed the available memory capacity, which also illustrates the importance of scenario selection.

\begin{table*}[t]
\centering
\caption{The cost comparison between different methods for eight forecasting settings (USD)}
\resizebox{1\linewidth}{!}{
\begin{tabular}{ccccccccc}
\hline
& Parametric & Non-parametric & VAE(Separate) & VAE(Joint) & GAN(Separate) & GAN(Joint) & Diff(Separate) & Diff(Joint) \\ \hline
Ground Truth         & 311305.18  & 311305.18      & 311305.18     & 311305.18  & 311305.18     & 311305.18  & 311305.18      & 311305.18   \\
Deterministic        & 424163.44  & 425584.03      & 421569.38     & 451049.72  & 398893.63     & 404846.03  & 411669.50      & 412576.34   \\
AO                   & 379128.07  & 379875.86      & 374705.55     & 355684.78  & 372536.22     & 354609.88  & 367380.31      & 360903.20   \\
DF                   & 379176.95  & 379314.22      & 372933.98     & 351629.16  & 371985.70     & 354189.79  & 366766.40      & 357832.51   \\
DF(Selector-KL)        & 378735.66  & 379047.19      & 370978.25     & 349299.50  & 371051.50     & 352048.19  & 366030.00      & 354384.19   \\
DF(Selector-Inner)     & 378795.16  & 379018.97      & 371471.41     & 349609.97  & 371537.03     & 352333.94  & 366652.59      & 354880.78   \\
DF(Selector-Entropy)   & 378796.00  & 379004.00      & 371783.44     & 349589.56  & 371022.63     & 351980.78  & 366416.94      & 354602.22   \\
\hline
\multirow{3}{*}{\begin{tabular}[c]{@{}c@{}}DF(Selector)-AO(Random)\\\hline Ground Truth\end{tabular}}
             & 0.13\%     & 0.27\%         & 1.20\%        & 2.05\%     & 0.48\%        & 0.82\%     & 0.43\%         & 2.09\%      \\
                     & 0.11\%     & 0.28\%         & 1.04\%        & 1.95\%     & 0.32\%        & 0.73\%     & 0.23\%         & 1.93\%      \\
                     & 0.11\%     & 0.28\%         & 0.94\%        & 1.96\%     & 0.49\%        & 0.84\%     & 0.31\%         & 2.02\%     \\
\hline              
\end{tabular}
}
\end{table*}

\begin{table*}[t]
\centering
\caption{Out-of-sample cost comparison of scenario selection methods based on the decision-focused generative model under eight forecasting settings (USD)}
\resizebox{1\linewidth}{!}{
\begin{tabular}{ccccccccc}
\hline
               & Parametric         & Non-parametric     & VAE(Separate)      & VAE(Joint)         & GAN(Separate)      & GAN(Joint)         & Diff(Separate)     & Diff(Joint)        \\ \hline
Random         & 379176.95          & 379314.22          & 372933.98          & 351629.16          & 371985.70          & 354189.79          & 366766.40      & 357832.51          \\
K-means        & 380402.78          & 380288.69          & 372293.81          & 350304.59          & 372014.31          & 354015.94          & 366290.13\,\textsuperscript{(3)}      & 357147.19          \\
K-medoids      & 382379.81          & 381859.66          & 374837.53          & 352800.22          & 374184.78          & 360248.97          & 369264.06      & 363306.59          \\
Hierarchical   & 380578.81          & 380528.47          & 372222.59          & 350236.97          & 371960.47          & 353812.00          & 366084.22\,\textsuperscript{(2)}      & 356602.53          \\
Selector-KL      & 378735.66\,\textsuperscript{(1)}          & 379047.19\,\textsuperscript{(3)}          & 370978.25\,\textsuperscript{(1)}          & 349299.50\,\textsuperscript{(1)}          & 371051.50\,\textsuperscript{(2)}         & 352048.19\,\textsuperscript{(2)}         & 366030.00\,\textsuperscript{(1)}     & 354384.19\,\textsuperscript{(1)}         \\
Selector-Inner   & 378795.16\,\textsuperscript{(2)}         & 379018.97\,\textsuperscript{(2)}            & 371471.41\,\textsuperscript{(2)}            & 349609.97\,\textsuperscript{(3)}            & 371537.03\,\textsuperscript{(3)}            & 352333.94\,\textsuperscript{(3)}            & 366652.59        & 354880.78\,\textsuperscript{(2)}            \\
Selector-Entropy & 378796.00\,\textsuperscript{(3)}          & 379004.00\,\textsuperscript{(1)}           & 371783.44\,\textsuperscript{(3)}           & 349589.56\,\textsuperscript{(2)}           & 371022.63\,\textsuperscript{(1)}           & 351980.78\,\textsuperscript{(1)}           & 366416.94      & 354602.22\,\textsuperscript{(3)}           \\ \hline
Selector-Average & \textbf{378775.60} & \textbf{379023.39} & \textbf{371411.03} & \textbf{349499.68} & \textbf{371203.72} & \textbf{352120.97} & 366366.51      & \textbf{354622.40}
\\
\hline
\end{tabular}
}
\label{table:cosr_comparison}
\end{table*}

\section{Conclusion}
\label{conclusion}
This work proposes a unified decision-focused scenario generation framework that captures the joint distribution of uncertainty quantities, such as loads and renewable energy generations, across buses and can be implemented with mainstream generative models, including VAE, GAN, and diffusion models. The results demonstrate that modeling the joint distribution not only improves marginal forecasting accuracy but also reduces downstream operational costs. Moreover, in contrast to existing scenario selection methods based on statistical similarity metrics, whose performance may vary across different models and settings, the proposed scenario selector achieves consistently lower operating costs across different model architectures and experimental settings.

\appendices
\ifCLASSOPTIONcaptionsoff
  \newpage
\fi
\bibliographystyle{IEEEtran}
\bibliography{ref.bib}

\end{document}